\documentclass{article}

\usepackage{url}

\usepackage{color}
\newcommand{\dd}{\mathop{}\mathopen{}\mathrm{d}}
\usepackage{amsmath}
\usepackage{amsthm}
\usepackage{bbm}
\usepackage{natbib}
\usepackage{graphicx}
\usepackage{amsmath}
\usepackage{amssymb}
\usepackage{amsthm}
\numberwithin{equation}{section}

\usepackage[a4paper, total={6in, 8in}]{geometry}

\newtheorem{remark}{Remark}[section]
\newtheorem{theorem}{Theorem}[section]
\newtheorem{proposition}[theorem]{Proposition}
\newtheorem{corollary}{Corollary}[section]

\begin{document}

\begin{center}
		\huge
		Explaining machine learning models using entropic variable projection
\end{center}

\begin{center}
 Fran\c{c}ois Bachoc$^*$\\[2pt]
Institut de Math\'ematiques de Toulouse\\
$^*${Corresponding author: francois.bachoc@math.univ-toulouse.fr}\\[2pt]
 Fabrice Gamboa\\[2pt]
Institut de Math\'ematiques de Toulouse \\
Artificial and Natural Intelligence Toulouse Institute (3IA ANITI) \\
fabrice.gamboa@math.univ-toulouse.fr\\[6pt]
Max Halford\\[2pt]
Institut de recherche en informatique de Toulouse \\
max.halford@irit.fr\\[6pt]
Jean-Michel Loubes\\[2pt]
Institut de Math\'ematiques de Toulouse \\
Artificial and Natural Intelligence Toulouse Institute (3IA ANITI) \\
{jean-michel.loubes@math.univ-toulouse.fr}\\[6pt]
and\\[6pt]
Laurent Risser\\[2pt]
Institut de Math\'ematiques de Toulouse \\
Artificial and Natural Intelligence Toulouse Institute (3IA ANITI) \\
laurent.risser@math.univ-toulouse.fr
\end{center}


\begin{abstract}
{In this paper, we present a new explainability formalism designed to \textcolor{black}{shed light on} how each input variable of a test set impacts the predictions of machine learning models. Hence, we propose a group explainability formalism for trained machine learning decision rules, based on their response to the variability of the input variables distribution.
	In order to emphasize the impact of each input variable, this formalism uses an information theory framework that quantifies the influence of all input-output observations based on entropic projections. This is thus the first unified and model agnostic formalism enabling data scientists to interpret the dependence between the input variables, their impact on the prediction errors, and their influence on the output predictions. Convergence rates of the entropic projections are provided in the large sample case. Most importantly, we prove that computing an explanation in our framework has a low algorithmic complexity, making it scalable to real-life large datasets. We illustrate our strategy by explaining complex decision rules learned by using XGBoost, Random Forest or Deep Neural Network classifiers on various datasets such as \textit{Adult Income}, \textit{MNIST}, \textit{CelebA}, \textcolor{black}{\textit{Boston Housing}, \textit{Iris}, as well as synthetic ones}. We finally make clear its differences with the explainability  strategies \textit{LIME} and \textit{SHAP}, that are based on single observations. Results can be reproduced by using the freely distributed Python toolbox \url{https://gems-ai.aniti.fr/}.} 
\\
{\bf 2000 Math Subject Classification:} 	94A15, 62B99.
\end{abstract}

\section{Introduction}
\label{sec:Introduction}

\subsection{\textcolor{black}{The need for explainability in machine learning}}

Machine learning algorithms build predictive models which are nowadays used for a large variety of tasks.
Over the last decades, the complexity of such algorithms has grown, going from simple and interpretable prediction models based on regression rules to very complex models such as random forests, gradient boosting, and deep neural networks. We refer to \cite{trevor2009elements} for a description of these methods.
Such models are designed to maximize the accuracy of their predictions at the expense of the interpretability of the decision rule.
Little is also known about how the information is processed in order to obtain a prediction, which explains why such models are widely considered as black boxes.

This lack of interpretability gives rise to several issues.
When an empirical risk is minimized, the training step may be unstable or highly dependent on the optimization procedure due for example to non-convexity and multimodality.
Another subtle, though critical, issue is that the optimal decision rules learned by a machine learning algorithm highly depend on the properties of the learning sample. If a learning sample presents unwanted trends or a bias, then the learned decision rules will reproduce these trends or bias, even if there is no intention of doing so.

In order to regulate the use of machine learning tools (and more generally artificial intelligence) in industry and society, different institutions have started writing laws which explicitly deal with interpretability. For instance, the European Parliament adopted in 2018 a law called GDPR (General Data Protection Regulation) to protect the citizens from decisions made without the possibility of explaining why they were taken, introducing a right for explanation in the civil code. In 2021, the European Parliament also proposed the Artificial Intelligence Act\footnote{\tiny\url{https://digital-strategy.ec.europa.eu/en/library/proposal-regulation-laying-down-harmonised-rules-artificial-intelligence}}, which lays down harmonized rules for the use of artificial intelligence in the European Union. The \textit{Article 13} of this act constrains high-risk AI systems to be \textit{sufficiently transparent to enable users to interpret the system's output and use it appropriately}. Hence, building intelligible tools to interpret the decisions made by black-box prediction models with desirable properties, such as those of deep neural-networks, is nowadays a key challenge for data scientists.

\subsection{\textcolor{black}{Existing work on explainability}} \label{ssec:existing:work}

Different methods have been proposed to make understandable the reasons leading to a prediction.
We mention early works by \citet{herlocker2000explaining} for recommender systems, \citet{DBLP:conf/nips/CravenS95} for neural networks and \citet{dzindolet2003role} or \citet{Lou:2012:IMC:2339530.2339556} for generalized additive models.
Another generic solution, described in \citet{baehrens2010explain} and \citet{Caruana:2015:IMH:2783258.2788613}, focused on medical applications.
In \citet{LiptonICML2016}, a discussion was opened to refine  the discourse on interpretability. Recently, a special attention has also  been given to deep neural systems.  We refer for instance to \citet{Hooker2019ABF,montavon2017methods,selvaraju2016grad} and references therein.
Clues for real-world applications are given in \citet{hall2018practical}. \textcolor{black}{Note that} sparse models (see \citet{buhlmann2011introduction} for a general introduction on the importance of sparsity) enable to identify important variables.

\textcolor{black}{It is important to distinguish between two types of explainability in machine learning. \textit{Local explainability} means that a single individual in a data base is investigated. For instance, in binary classification, typical findings can be the two following: (1) A specific variable is the main cause of the individual's predicted binary label taking a given value (2) A specific variable can be modified in order to change the individual's predicted binary label. \textit{Global explainability} means that an entire data base is investigated. A typical finding can be: on average, increasing a specific variable increases the proportion of individuals for which the predicted binary label is the first one.}

\textcolor{black}{Nowadays, the two main methods that are used in practice to explain algorithmic decisions are LIME \citep{ribeiro2016should} and SHAP \citep{shap} (local explainability). In LIME, it is}  proposed to locally approximate  a black-box model  by a linear model. This linear surrogate mimics locally the behaviour of the algorithm. Hence the analysis is performed to understand the feature importance of the variables used in the linear model and thus to explain the decisions of the original model. \textcolor{black}{Each indiviual corresponds to a different local linear model and thus to a different interpretation of the variables' importance, indeed making LIME a local method.} SHAP consists in using so-called Shapley indices that provide a ranking of the importance of the different variables in the model outputs. 

\textcolor{black}{Local explainability methods differ from our proposal (see Section \ref{ssec:our:method:intro}), which aims at global explainability. We note that these two objectives are complementary. Local explainability is justified for instance when dealing with human individuals, as then an individual would be interested only in understanding the cause of the machine learning decision (prediction) in their case, and would not be interested in understanding how the machine learning algorithm ``work'' for the entire population. In contrast, when one aims at understanding a machine learning algorithm, explaining each individual separately is not a solution (especially given the very large size of modern databases). Global explainability then provides manageable indicators for this.}

\textcolor{black}{In particular, the reference \cite{waldchen2021} considers a function $\phi : \{0,1\}^p \to \{0,1\}$ (a binary classifier operating on binary variables) and, for a given $x = (x_1,\ldots,x_p) \in \{0,1\}^p$, aims at finding a subset of $x$, of minimal size, such that with large probability, the output of $\phi$ is unchanged when the values of $x$ on the complement of the subset are randomly sampled. Hence, \cite{waldchen2021} address explainability, as the variables in the subset are arguably the most important ones to explain the decision (the value $\phi(x)$). Furthermore, \cite{waldchen2021} address local explainability, as the analysis is made for a specific $x$. Thus, our proposal and \cite{waldchen2021} address complementary objectives. 
}

\textcolor{black}{A category of methods explains which input variables are the most likely to change the output algorithmic decisions. These explainability methods more specifically imagine what would have changed in the model output if the inputs were changed. This kind of formalism has been referred to as counterfactual explanations. Generating this kind of explanations enables us to understand what causes the changes in the decisions. This requires knowing in advance how the data are generated or a causal graph that describes the causal relationships between the different variables. In many cases, such knowledge is unavailable. Hence, either the causal model has to be estimated, which is computationally a very demanding task in high dimension, or, otherwise inferring the counterfactuals has to be investigated.  }
Explanation methods based on this framework have already been  used in the literature. Counterfactual models \citep{WachterHJLT18} are proposed in \citet{Goyal_ICML2019} to explain how the predictions made by a classifier on a query image can be changed by transforming a region of this image.
In the same vein, a method called integrated gradients was specifically designed in \citet{MukundICML17} for the interpretability of single predictions using neural networks. This method provides local explainability.
In the \textit{Fair learning} community,  counterfactual models are also used to assess whether the predictions of machine learning models are fair \citep{NIPS2017_6995,FlipTest2020}.
An individual explanation method on images through adversarial examples was also presented in \citet{NIPS2019_9717}.
In \citet{KohMLR2017} the authors finally proposed a strategy to understand black-box models, in a parametric setting.
\textcolor{black}{Importantly for us, the validity of the previous methods may be questionable  since they rely on the creation of  counterfactual observations. Such counterfactuals may be out of distribution and may thus not reflect the desired behaviour of the observation points. In general, we believe that}  running the algorithm over observations which are created artificially by increasing stepwise the value of a particular variable is not a desirable solution for explainability. \textcolor{black}{
Indeed, as explained above the new individuals may actually be outliers. In addition, the correlations between variables are usually not taken into account when creating new individuals by changing one variable at a time. 
}

\textcolor{black}{
 We also point out that some existing methods highlight important variables but do not explain their role. For instance, the variables' importance is often computed using feature importance or Gini indices (see in~\citet{Raileanu2004} or \citet{trevor2009elements}). Yet such indices are computed without investigating the particular effects of each variable (for instance without indicating if they have a positive or negative impact on the output) and without explaining their particular role in the decision process. }

\textcolor{black}{The above shortcomings of existing methods motivate our proposed method below: a global explainability framework that does not need to create new individuals, but simply to study a real existing database by reweighting the individuals, as described next. In addition, the method not only highlights the importance of the variables, but also analyzes the role of important variables (for instance if a variable has a positive or negative effect on a quantity of interest).}

\subsection{\textcolor{black}{Our proposed method}} \label{ssec:our:method:intro}

\indent In this work, we specifically present a strategy to generate test distributions that present a controlled deviation  with respect to an original test distribution. \textcolor{black}{To achieve this, we first consider specific deformations that we want to apply to the machine learning test set. These deformations can be for instance an increased mean or variance of one input or output variable. It can alternatively be a decreased correlation between two variables, and even be a lower error rate on the predictions. Once the deformation chosen, it is enforced in the dataset while maintaining the distribution of the deformed dataset as close as possible to the original distribution.  The deformation can be seen as a {\bf stress} or a {\bf bias} applied to the original distribution. This therefore makes it possible to study how the algorithmic decisions of the stressed distribution differ from the original one, so the behaviour of the prediction algorithm can be explained. To compute the stressed distribution, we consider the framework of the  entropic projection method: when the deformation can be written as the mean of a chosen function, minimizing the  Kullback-Leibler divergence between the original  distribution and the new one, that satisfies the constraint, has a feasible solution. This solution is actually obtained by reweighting the empirical distribution of the original sample. This therefore enables to simply create new distributions of observations satisfying the chosen constraint, and to study how the algorithmic decisions are affected by the stress condition, which contributes to the analysis of the behaviour of the algorithm.
This not only explains} the role played by the input variables, but can also enable to infer causal properties when looking at changes of predictions caused by changes in the variables.

\textcolor{black}{The method addresses the two shortcomings discussed in Section \ref{ssec:existing:work}. First, it enables to answer the question {\it what would have happened if } the data had been modified in a certain way, but nevertheless ensures that the modified data are plausible, in the sense that their distribution still corresponds to the distribution for which the algorithm has been trained for. 
In fact, no new individual needs to be created, the existing ones being simply reweighted. In line with this, no additional calls to the machine learning algorithm are necessary. 
Second, not only does the method points out which variables are important, but it also characterizes this importance, for instance, in terms of a positive or negative impact on the prediction or the prediction error. Furthermore, the method is able to detect complex importance features, in particular it can discriminate between correlation-importance and causation-importance (Section \ref{sssec:SimulatedData:corr:caus}).}

Besides introducing this novel framework in explainable machine learning, we prove the optimality of the reweighting methodology and obtain a theoretical control on the generated distribution with respect to Wasserstein distance. Furthermore, the method is scalable to large datasets.

We remark that enforcing a bias to the original distribution is inspired by the fields of sensitivity analysis and computer experiments \citep{lemaitre2015density,saltelli2008global}, and also that the subsequent reference \cite{de2021transport} adapts our procedure to the Wasserstein distance.

The paper falls into the following parts. Methodology is explained in Sections~\ref{sec:Methodology} and \ref{s:main2}. \textcolor{black}{The content of these two sections mainly consists in our original contributions, see in particular Remark \ref{remark:summary:novelty:secdeux} for Section \ref{sec:Methodology}.}
Results are given in Section~\ref{res:results} and the discussions are finally developed in Section~\ref{sec:discussion}. An appendix contains the proofs.

\section{Optimal perturbation of machine learning datasets}
\label{sec:Methodology}



We consider a test set $\{(X_i,Y_i)\}_{i = 1,\dots, n}$, where $X_i=(X^1_i,\ldots,X^p_i)$ is an observed input vector while $Y_i$ is the corresponding true output, on which we consider the outcome of black box decision rules $f: \mathbb{R}^p \to \mathbb{R}$. We consider throughout this paper that $f$ has been learned from a training set and is fixed. We set  $\hat{Y_i} = f(X_i)$, the predicted output. Hence we have at hand values $(X_i,\hat{Y}_i,Y_i)$ for $i=1,\dots,n$.

Our goal is to explain the global behaviour of $f$. To achieve this, we propose to study the response of $f$ to distributional perturbations (stress or bias) of the input variables. Since $f$ has been learnt using data following a given distribution, the domain of validity of the algorithm should not deviate too much from this initial distribution. Hence, we propose to build perturbed distributions that are as close as possible to the initial distribution using an information theory method. We will show that this amounts to reweighing the observations by proper weights calibrated to incorporate the chosen perturbation on the input variables  as explained in Section~\ref{s:impact:mean}. The methodology to make this problem well posed and to quickly compute the optimal weights is the core of this paper.

\begin{remark} \label{rem:test:set}
\textcolor{black}{
Often,
the dataset $\{(X_i,Y_i)\}_{i = 1,\dots, n}$ is directly available from the machine learning problem at hand (for instance the \textit{Boston Housing} dataset of Section \ref{ss:regressionCase:appli}). Then as seen above, the machine learning predictor can be called to obtain the $\hat{Y}_i$'s. 
In can also be the case that only the trained machine learning predictor function $f : \mathbb{R}^p \to \mathbb{R}$ is available. In this case, as $X_1 , \ldots , X_n$ are not available, we would not aim at interpreting the distribution of the input variables of the machine learning model. We would then typically choose a ``neutral'' distribution for them, for instance uniform on $[0,1]^p$ (after suitable componentwise scalings), then sample $X_1,\ldots,X_n$ from it and then provide explainability tools on $\{(X_i,\hat{Y}_i)\}_{i = 1,\dots, n}$. Indeed, some of the provided explainability tools below do not necessitate to have true values $Y_1,\ldots,Y_n$ (for instance we would interpret the predictions of the machine learning algorithm but not its errors).} 
\end{remark}

\subsection{General optimal perturbations under moment constraints in machine learning} \label{s:stress}

In order to experiment and to explore the behavior of a predictive model, a natural idea is to study its response to biased inputs.
Different ways are possible to create modifications of \textcolor{black}{the probability measure of the test set}
\begin{equation} \label{eq:Qn}
Q_{n} = \frac{1}{n}  \sum_{i=1}^n  \delta_{X_i,\hat{Y}_i,Y_i}.
\end{equation}

\textcolor{black}{The above notation implies in particular that for any function $g : \mathbb{R}^{p+2} \to \mathbb{R}$, its expectation with respect to $Q_{n}$ is computed with the well-known formula \begin{equation} \label{eq:expectation:Qn}
\frac{1}{n}  \sum_{i=1}^n  g(X_i,\hat{Y}_i,Y_i),
\end{equation}
and can be interpreted as providing the same weight $1/n$ to all data points $(X_i,\hat{Y}_i,Y_i)$ in the test set. 
}

In this paper, we consider an information theory framework in which we modify the distribution  of the original test set $Q_{n} = \frac{1}{n}  \sum_{i=1}^n  \delta_{X_i,\hat{Y}_i,Y_i}$,
by stressing the mean value of a function $\Phi$ (or simply by stressing the mean value of variables).
We minimize the Kullback-Leibler information (also called mutual entropy) with respect to $Q_n$, making the problem well posed.

First, let us recall the definition of the Kulback-Leibler information.  Let $(E,\mathcal{B}(E))$ be a measurable space and $Q$ a probability measure on $E$. If $P$ is another probability measure on $(E,\mathcal{B}(E))$, then the Kullback-Leibler information $\mathrm{KL}(P,Q)$ is equal to $\int_{E}\log\frac{\dd P}{\dd Q}\dd P$, if  $P\ll Q$ and
$\log\frac{\dd P}{\dd Q}\in L^1(P)$, and $+\infty$ otherwise.

For a given $k \geq 1$, let
\[
\Phi: (X,\hat{Y},Y) \in \mathbb{R}^{p+2} \mapsto \Phi(X,\hat{Y},Y) \in \mathbb{R}^k
\]
be a measurable function representing the shape of the stress deformation on the whole input (for instance $\Phi$ can be a selected input coordinate when $k=1$, see Section \ref{s:impact:mean}). Note that our results are stated for a generic function $\Phi$ of all variables $(X,\hat{Y},Y)$. Of course, this includes the case of functions depending only on $(X,Y)$ \textcolor{black}{or $(X,\hat{Y})$}. \\
\textcolor{black}{ The problem can be stated as follows: given the distribution $Q_n$, we aim to construct a distribution close to $Q_n$ but satisfying a constraint expressed through the mean of the chosen function $\Phi$. Actually, } for $t \in \mathbb{R}^k$, we  aim at finding a new distribution $Q_t$ satisfying the constraint
\[
\int_{\mathbb{R}^k} \Phi(x) \dd Q_t(x) = t,
\]
and being the closest possible to the initial empirical distribution $Q_n$ in the sense of Kullback-Leibler divergence, \textit{i.e.}  with $\mathrm{KL}(Q_t,Q_n)$ as small as possible. \textcolor{black}{Note that the above display means that $t$ is the expectation $\mathbb{E}(\Phi(Z))$ for any random variable $Z$ with distribution $Q_t$.} \\

We set for two vectors $x,y \in \mathbb{R}^k$ the scalar product as $\langle x,y \rangle=x^\top y$. We next characterize the new distribution $Q_t$.

\begin{theorem} \label{thm:discrete:reweigth:multidim}
	Let $t \in \mathbb{R}^k$ and $\Phi : \mathbb{R}^{p+2} \to \mathbb{R}^k$ be measurable.
	Assume that $t$ can be written as a convex combination of $\Phi(X_1,\hat{Y}_1,Y_1) , \ldots , \Phi(X_n,\hat{Y}_n,Y_n)$, with positive weights. Assume also that the empirical covariance matrix
	$\mathbb{E}_{Q_n} (\Phi \Phi^\top)  - \mathbb{E}_{Q_n} (\Phi ) \mathbb{E}_{Q_n} (\Phi^\top)$ is invertible.

	Let $\mathbb{P}_{\Phi,t}$ be the set of all probability measures $P$ on $\mathbb{R}^{p+2}$ such that $
	\int_{\mathbb{R}^{p+2}} \Phi(x) \dd P(x)=t$.
	For a  vector $\xi \in \mathbb{R}^k$, let $Z(\xi):=\frac{1}{n} \sum_{i=1}^n e^{\langle  \Phi(X_i,\hat{Y}_i,Y_i) , \xi \rangle}$. Define now $\xi(t)$ as the unique minimizer of the strictly convex function
	$H(\xi):=\log Z(\xi)-\langle\xi,t\rangle$.
	Then,
	\begin{equation}
		Q_t:=\mathrm{arginf}_{P\in\mathbb{P}_{\Phi,t}} \mathrm{KL}(P,Q_n) \label{thesol:Qn}
	\end{equation}
	exists and is unique. Furthermore, we have
	\begin{equation}
		Q_t=
		\frac{1}{n}
		\sum_{i=1}^n
		\lambda_i^{(t)}
		\delta_{X_i , \hat{Y}_i , Y_i},
		\label{gibbs}
	\end{equation}
	with, for $i=1,\ldots,n$,
	\begin{equation} \lambda^{(t)}_i
		= \exp
		\bigg( \langle\xi(t) ,  \Phi(X_i,\hat{Y_i},Y_i) \rangle - \log Z(\xi(t))
		\bigg).  \label{defi:change:weight:multi:dim}
	\end{equation}
\end{theorem}

\textcolor{black}{It is instructive to compare \eqref{gibbs} and \eqref{eq:Qn}. For a function $g : \mathbb{R}^{p+2} \to \mathbb{R}$, its expectation with respect to $Q_t$ in \eqref{gibbs} is
\[
	\frac{1}{n}
		\sum_{i=1}^n
		\lambda_i^{(t)} g(X_i , \hat{Y}_i , Y_i ).
\]
Compared to \eqref{eq:expectation:Qn}, the same values $g(X_i , \hat{Y}_i , Y_i )$ are used, that is the two distributions $Q_n$ and $Q_t$ have the same support, but some data points in the above display have more weight that others in computing the expectation. For instance, when $k=1$ (that is $\Phi $ is univariate) and $t > \int_{\mathbb{R}^{p+2}} \Phi(x) \dd Q_n(x)$ (the mean of $\Phi$ is increased), then \eqref{defi:change:weight:multi:dim} shows that data points with larger values of $\Phi(X_i , \hat{Y}_i , Y_i)$ will have larger weights. Hence, we can favour observations with large values of $\Phi$, which is the principle of stressing means for variable importance in Theorem \ref{th:mainenpractice}.
}

 A particularly appealing aspect of Theorem \ref{thm:discrete:reweigth:multidim}, \textcolor{black}{as just discussed,} is that $Q_{t}$ is supported by the same observations as $Q_n$, the mean change for $\Phi$ only leading to different weights for the observations.   Hence, sampling  new stressed test sets  does not require to create new input-output observations $(X_i,\hat{Y}_i,Y_i)$ but only to  compute the weights $\lambda_i^{(t)}$. This can be solved very quickly using \eqref{defi:change:weight:multi:dim}.
 This desirable property is obtained through the choice of the Kullback-Leibler information as a measure of similarity between $Q_t$ and $Q_n$. Theorem \ref{thm:discrete:reweigth:multidim} also yields a favorable optimization problem, both theoretically and numerically, as discussed in Section \ref{subsection:optim:practical}.

 Set
	\[
	t_0 = \int_{\mathbb{R}^k} \Phi(x) \dd Q_n(x) = \frac{1}{n} \sum_{i=1}^n \Phi(X_i,\hat{Y}_i,Y_i),
	\]
	the empirical mean of $\Phi$ with the distribution $Q_n$. The quantity $t - t_0$ can be thus understood as the amount of change on the mean of $\Phi$ induced by changing  $Q_n$ into $Q_t$.   We remark that, in Theorem \ref{thm:discrete:reweigth:multidim}, the condition that $t$ can be written as a convex combination of $\Phi(X_1,\hat{Y}_1,Y_1) , \ldots , \Phi(X_n,\hat{Y}_n,Y_n)$, with positive weights, is almost minimal. Indeed, it is necessary that $t$ can be written as a convex combination of $\Phi(X_1,\hat{Y}_1,Y_1) , \ldots , \Phi(X_n,\hat{Y}_n,Y_n)$ (otherwise the set of distributions that are mutually absolutely continuous to $Q_n$ and yield expectation $t$ for $\Phi$ is empty). For all considered examples in this paper, this condition in Theorem \ref{thm:discrete:reweigth:multidim} was not restrictive.

\subsection{Problem resolution using gradient descent} \label{subsection:optim:practical}

Let us denote $v_i = (X_i,\hat{Y_i},Y_i)$ for $i=1,\ldots,n$. In Theorem \ref{thm:discrete:reweigth:multidim} we want to minimize, over $\xi \in \mathbb{R}^k$,
\begin{equation}\label{eq:minimizedFctMultiDim}
	H(\xi) =
	\log{\left(\frac{1}{n} \sum_{i=1}^n \exp{( \langle \xi, \Phi{(v_i)}\rangle)}  \right)} - \langle\xi  , t \rangle.
\end{equation}
It can be seen that the optimization problem is convex and can therefore be addressed very efficiently using gradient descent.
The gradient of \eqref{eq:minimizedFctMultiDim} is:
\begin{equation}
	\nabla_{\xi} H(\xi) =
	\frac{
		\sum_{i=1}^n \Phi{(v_i)} \exp{ \langle \xi ,  \Phi(v_i) \rangle }
	}{
		\sum_{i=1}^n \exp{ \langle \xi ,  \Phi(v_i) \rangle }
	}
	- t \,.
\end{equation}
Once the optimized $\xi(t)$ is reached, the weights $\lambda^{(t)}_i$, $i=1,\ldots,n$, associated to the observations are given in \eqref{defi:change:weight:multi:dim}.

Our framework makes it possible to deal with very large databases without computing any new machine learning prediction. This differs from existing techniques based on perturbed observations as \textit{e.g.}  in  LIME \citep{ribeiro2016should},  where the data used for testing are created by changing randomly the labels or by bootstrapping the observations.

\subsection{Asymptotic rate of convergence} \label{ssec:asymptotic:consistency}

While, in this paper, the test set $(X_i,\hat{Y}_i,Y_i)_{i=1,\ldots,n}$ with empirical distribution $Q_n$ is considered fixed, in Section \ref{ssec:asymptotic:consistency} (and only in Section \ref{ssec:asymptotic:consistency}), we assume that this test set is random, composed of i.i.d. realizations of a distribution $Q^{\star}$.

The following proposition proves a statistical rate of convergence of our methodology. We show that the optimally perturbed distribution $Q_t$ of Theorem \ref{thm:discrete:reweigth:multidim} (defined w.r.t. $Q_n$, $\Phi$ and $t$) converges to the corresponding optimally perturbed distribution $Q^{\star}_t$, (defined w.r.t. $Q^{\star}$, $\Phi$ and $t$). The convergence is measured by the $L^1$ Wasserstein distance $\mathcal{W}_1$, defined by, for two distributions $P,Q$ on $\mathbb{R}^{p+2}$ with finite first moments,
\[
\mathcal{W}_1(P,Q)
=
\inf_{U \sim P , V \sim Q}
\mathbb{E}\left(
||U - V||
\right),
\]
where the above infimum is taken over all pairs of (dependent or independent) random variables $U,V$ such that $U \sim P , V \sim Q$. We refer for instance to \citet{peyre2019computational} for the definition and computation of $\mathcal{W}_1$ and to \citet{bachoc2017gaussian,cazelles18geodesic} for recent work related to it.

\begin{proposition} \label{prop:rate:convergence}
	Let $\Phi: \mathbb{R}^{p+2} \to \mathbb{R}^k$ and $t \in \mathbb{R}^k$ be fixed.
	Assume that the support of $Q^{\star}$ is bounded and that $\Phi$ is bounded in absolute value and Lipschitz continuous on the support of $Q^{\star}$.
	Assume also that for $v \in \mathbb{R}^k$, $b \in \mathbb{R}$, $Q^{\star} (\{x \in \mathbb{R}^{p+2} ;  \langle v , \Phi(x) \rangle = b \}) = 1$ if and only if $v = 0$ and $b=0$. Assume finally that there exists a distribution $Q$, mutually absolutely continuous to $Q^{\star}$, such that $\int_{\mathbb{R}^{p+2}} \Phi(x) \dd Q (x) = t$.

	Then there exists a unique measure $Q^{\star}_t$ on $\mathbb{R}^{p+2}$ such that $\int_{\mathbb{R}^{p+2}} \Phi(x) \dd Q^{\star}_t(x)  = t$ and $\mathrm{KL}(Q^\star_t,Q^\star)$ is minimal.
	Furthermore, as $n \to \infty$, with $Q_t$ given in Theorem \ref{thm:discrete:reweigth:multidim},
	\[
	\mathcal{W}_1
	\left(
	Q_t,Q^{\star}_t
	\right)
	=
	O_p \left( n^{- 1/(p+2)} \right).
	\]
\end{proposition}

The rate of convergence $O_p \left( n^{- 1/(p+2)} \right)$ is standard for the $L^1$ Wasserstein distance in dimension $p+2$, see \citet{fournier2015rate} for instance. The proof of Proposition \ref{prop:rate:convergence} combines techniques from the analysis of Wasserstein distances and from M-estimation with convex objective functions.

For the sake of conciseness, Proposition \ref{prop:rate:convergence} is stated under boundedness assumptions and with independent realizations from $Q^\star$. These conditions could be weakened.

\textcolor{black}{The previous proposition proves the consistency of the method to generate new distributions that can be used as different {\bf stress} tests to understand the modification of the outcome of the algorithm. Also, this proposition shows that if the available test set follows a distribution of interest $Q^{\star}$, then having access to the test set is sufficient to draw sound conclusions about $Q^{\star}$. In other words, our explainability framework can handle the behavior of the machine learning model on unseen data, provided these unseen data have a distribution close to that of the available test set. We remark that if unseen data have a distribution completely different from the test set, then our explanations computed on the test set will be less relevant to them. This is a specific problem in machine learning, that is called transfer learning \citep{pan2009survey}, and is outside of the scope of this work.
For instance, similarly, as discussed in Section \ref{ssec:existing:work}, \cite{waldchen2021} fix subsets of input variables (as small as possible) such that the probability that the output is changed, when sampling from the remaining variables, is small. The fixed variables are then considered to be important. This notion of importance depends on the distribution used to define this probability. Hence, similarly as in our case, providing interpretations based on a specific distribution of the variables (for instance learnt from the test set in our case) can be invalid for different distributions of these variables. 
}

\textcolor{black}{
Furthermore, our framework is versatile since it can be implemented for any  function $\Phi$ that models different types of deviations that can be incorporated.}
Hereafter, the following section is devoted to  different examples of functions  $\Phi$, which shed light on the impact of various features of the input variables.

\subsection{\textcolor{black}{Applications: stressing means for variable importance, variances and covariances}}
\label{s:impact:mean}

We now apply  Theorem~\ref{thm:discrete:reweigth:multidim} to the special case of perturbing the mean of one of the $p$ input variables,
meaning that $\Phi$ is valued in $\mathbb{R}$ (\textit{i.e.} $k=1$).

\begin{theorem} \label{th:mainenpractice}
	Let $t \in \mathbb{R}$ and $j_0 \in \{1 , \ldots , p\}$.
	Assume that $ \min_{i=1}^n X_i^{j_0} < t < \max_{i=1}^n X_i^{j_0}$.

	Let $\mathbb{P}_{j_0,t}$ be the set of probability measures on $\mathbb{R}^{p+2}$ such that, when $(X,\hat{Y},Y)$ follows a distribution $P \in \mathbb{P}_{j_0,t}$, we have $  \mathbb{E} ({X}^{j_0})   =  t$.
	For $\xi \in \mathbb{R}$, let $Z(\xi):=\frac{1}{n} \sum_{i=1}^n e^{ \xi X_i^{j_0} }$. Define now $\xi(t)$ as the unique minimizer of the strictly convex function
	$H(\xi):=\log Z(\xi)- \xi t$.
	Then,
	\begin{equation*}
		Q_{j_0,t}:=\mathrm{arginf}_{P \in \mathbb{P}_{j_0,t}} \mathrm{KL}(P,Q_n)
	\end{equation*}
	exists and is unique. Furthermore, we have
	\begin{equation*}
		Q_{j_0,t}=
		\frac{1}{n}
		\sum_{i=1}^n
		\lambda_i^{(j_0,t)}
		\delta_{X_i , \hat{Y}_i , Y_i},
	\end{equation*}

	with, for $i=1,\ldots,n$,
	\begin{equation*}
		\lambda_i^{(j_0,t)}
		= \exp
		\bigg( \xi(t) X_i^{j_0} - \log Z(\xi(t))
		\bigg).
	\end{equation*}
\end{theorem}
This theorem enables to re-weight the observations of a variable so that its mean increases or decreases. This is then used in Section~\ref{s:main2} to understand the particular role played by each variable.
\textcolor{black}{To give a concrete example, in Section \ref{res:resultsTwoClass} we address the \textit{Adult Income} dataset, with $p=14$ and where each $X_i \in \mathbb{R}^{14}$ provides characteristics of a given adult individual. We are then interested in the first component $X_i^{j_0}$, $j_0=1$, representing the age.
The average age in the dataset is approximately $38.6$.
With Theorem \ref{th:mainenpractice}, we can increase this average age to, say, 45, by weighting the individuals with weights that are increasing with age. On the \textit{Adult Income} dataset, the outputs $Y_i$ and $\hat{Y}_i$ (its prediction by random forest on the right-hand side of Figure \ref{fig:ResultsAdultIncome}) are binary, with 1 indicating higher income and 0 indicating lower income. 
We can see that the  the portion of 1 is increased with the weights corresponding to average age 45, from which we conclude that age has a positive impact on having higher income.}

\vskip .1in


The next corollary enables to stress the means of several variables at the same time.

\begin{corollary}[perturbing several means] \label{cor:stress:means}
	Let $1 \leq c \leq p$ and let $j_1,\ldots,j_c$ be two-by-two distinct in $\{ 1 , \ldots , p\}$. Let $t_1,\ldots,t_c \in \mathbb{R}$.
	Assume that there exists a convex combination of $(X_i^{j_1},\ldots,X_i^{j_c})_{i=1,\ldots,n}$ with positive weights that is equal to $(t_1,\ldots,t_c)$. Assume also that the empirical covariance matrix of $ (X_i^{j_1},\ldots,X_i^{j_c})_{i=1,\ldots,n} $ is invertible.
	Then, there exists a unique distribution $Q_{t_1,\ldots,t_c}$ on $\mathbb{R}^{p+2}$ such that for $(X^1,\ldots,X^p,\hat{Y},Y) \sim Q_{t_1,\ldots,t_c}$ we have $\mathbb{E} (X^{j_a}) = t_a$ for $a = 1,\ldots,c$ and such that $\mathrm{KL}(Q_{t_1,\ldots,t_c},Q_n)$ is minimal. This distribution is obtained by the distribution $Q_t$ in Theorem \ref{thm:discrete:reweigth:multidim}, in the case where $k=c$ and $\Phi(X^1,\ldots,X^p,\hat{Y},Y) = (X^{j_1},\ldots,X^{j_c})$.
\end{corollary}

The next corollary enables to stress the variance of a variable while preserving its mean $m_{j_0} = \frac{1}{n}\sum_{i=1}^n X^{j_0}_i$.

\begin{corollary}[perturbing the dispersion] \label{cor:stress:variance}
	Let $j_0 \in \{ 1 , \ldots , p\}$. Let $v \in [0 , \infty)$.
	Assume that there exists a convex combination of $(X_i^{j_0},(X_i^{j_0})^2)_{i=1,\ldots,n}$ with positive weights that is equal to $(m_{j_0}, m_{j_0}^2 + v )$. Assume also that the empirical covariance matrix of $ (X_i^{j_0},(X_i^{j_0})^2)_{i=1,\ldots,n} $ is invertible.
	Then, there exists a unique distribution $Q_{j_0,v}$ on $\mathbb{R}^{p+2}$ such that for $(X^1,\ldots,X^p,\hat{Y},Y) \sim Q_{j_0,v}$ we have $\mathbb{E} (X^{j_0}) = m_{j_0}$ and $\mathrm{Var}(X^{j_0}) = v$ and such that $\mathrm{KL}(Q_{j_0,v},Q_n)$ is minimal. This distribution is obtained by the distribution $Q_t$ in Theorem \ref{thm:discrete:reweigth:multidim}, in the case where $k=2$, $\Phi(X^1,\ldots,X^p,\hat{Y},Y) = (X^{j_0},(X^{j_0})^2)$ and $t = (m_{j_0}, m_{j_0}^2 + v )$.
\end{corollary}

Finally, we next show how to stress the covariance between two variables while preserving their means $m_{j_1} = \frac{1}{n}\sum_{i=1}^n X^{j_1}_i$ and $m_{j_2} = \frac{1}{n}\sum_{i=1}^n X^{j_2}_i$.

\begin{corollary}[perturbing the covariance] \label{cor:stress:covariance}
	Let $j_1 , j_2 \in \{ 1 , \ldots , p\}$ be distinct. Let $c \in \mathbb{R}$.
	Assume that there exists a convex combination of $(X_i^{j_1},X_i^{j_2},X_i^{j_1}X_i^{j_2})_{i=1,\ldots,n}$ with positive weights that is equal to $(m_{j_1}, m_{j_2} , m_{j_1} m_{j_2} + c )$.
	Assume also that the empirical covariance matrix of $ (X_i^{j_1},X_i^{j_2},X_i^{j_1}X_i^{j_2})_{i=1,\ldots,n} $ is invertible.
	Then, there exists a unique distribution $Q_{j_1,j_2,c}$ on $\mathbb{R}^{p+2}$ such that for $(X^1,\ldots,X^p,\hat{Y},Y) \sim Q_{j_1,j_2,c}$ we have $\mathbb{E} (X^{j_1}) = m_{j_1}$, $\mathbb{E} (X^{j_2}) = m_{j_2}$ and $\mathrm{Cov}(X^{j_1},X^{j_2}) = c$ and such that $\mathrm{KL}(Q_{j_1,j_2,c},Q_n)$ is minimal. This distribution is obtained by the distribution $Q_t$ in Theorem \ref{thm:discrete:reweigth:multidim}, in the case where $k=3$, $\Phi(X^1,\ldots,X^p,\hat{Y},Y) = (X^{j_1},X^{j_2},X^{j_1}X^{j_2})$ and $t = (m_{j_1}, m_{j_2} , m_{j_1} m_{j_2} + c )$.
\end{corollary}

\textcolor{black}{In Section \ref{sssec:SimulatedData:corr:caus}, we show that Corollary \ref{cor:stress:covariance} can be used to concretely discriminate between correlation and causation. Indeed, a variable $j_1$ can be correlated with the output of interest, which means that it helps to predict the output values, but can have no causal influence on the output, which means that changing the variable while keeping the other variables fixed has no impact on the output.
This can happen when the variable $j_1$ is correlated with another variable $j_2$ (that can be called a confounding variable) that itself has a causal influence.  
This can be detected by combining Theorem \ref{th:mainenpractice} and Corollary \ref{cor:stress:covariance}. We can use simply Theorem \ref{th:mainenpractice} to detect for instance that increasing the mean of the variable $j_1$ increases the mean of the output. Then we can impose a zero covariance between variables $j_1$ and $j_2$ with Corollary \ref{cor:stress:covariance} while increasing the mean of the variable $j_1$, to observe that this does not increase the mean of the output any longer.}

\begin{remark} \label{remark:summary:novelty:secdeux}
\textcolor{black}{Most of the content of Section \ref{sec:Methodology} is original. Section \ref{sec:Methodology}
exploits the known results from \citet{Csis1} and \citet{Csis2} (restated as Theorem \ref{th:main} in the appendix). Then Theorems \ref{thm:discrete:reweigth:multidim} and \ref{th:mainenpractice} and Corollaries \ref{cor:stress:means} to \ref{cor:stress:covariance} are original results which proofs use Theorem \ref{th:main}. Section \ref{subsection:optim:practical} provides an (original) gradient expression for Theorem \ref{thm:discrete:reweigth:multidim}, based on standard calculations. Finally Section \ref{ssec:asymptotic:consistency} provides an original theoretical result (Proposition \ref{prop:rate:convergence}) regarding rates of convergence. Its proof originally combines techniques from the analysis of Wasserstein distances and from M-estimation with convex objective functions.}
\end{remark}

\section{Explainable models for classification \textcolor{black}{and regression} using optimally perturbed datasets} \label{s:main2}

In this section we consider that the transformation $\Phi : \mathbb{R}^{p+2} \to \mathbb{R}^k$ and the target multidimensional moment $t \in \mathbb{R}^k$ have been selected and that the conditions of Theorem \ref{thm:discrete:reweigth:multidim} are satisfied. This theorem provides the optimally perturbed distribution $Q_t$, given by the weights $(\lambda_i^{(t)})_{i=1,\ldots,n}$. We now suggest various quantitative properties of $Q_t$ (that we call quantities of interest), that can quantify the behavior of the studied black box decision rule.

We shall focus on \textcolor{black}{three} classical situations encountered in machine learning: binary classification, multi-class classification \textcolor{black}{and regression}.

\subsection{The case of binary classification}  \label{ss:binary:classif}

Consider that $Y_i$ and $\hat{Y}_i=f(X_{i})$  belong to  $\{ 0 , 1\}$ for all $i = 1 \ldots , n$. This corresponds to the binary classification problem for which the usual loss function  is $\ell (Y,f(X))=\mathbbm{1}_{\left\{ Y \neq f(X) \right\}}.$   We suggest to use the indicators described hereafter for the perturbed distribution $Q_t = \frac{1}{n} \sum_{i=1}^n \lambda^{(t)}_i \delta_{(X_{i},\hat{Y}_i,Y_i)}$. Explaining the decision rules may first consist in quantifying the evolution of the error rate as a function of $t - t_0$, hence the first indicator is the error rate, \textit{i.e.}
\[
\mathrm{ER}_{t}
=
\frac{1}{n}
\sum_{i=1}^n
\lambda^{(t)}_i
\mathbbm{1}_{\left\{  f(X_i) \neq Y_{i} \right\}}.
\]
In terms of interpretation, when $\Phi$ is given as in Theorem \ref{th:mainenpractice}, with $\Phi(X^1,\ldots,X^p,\hat{Y},Y) = X^{j_0}$, $t$ corresponds to the new (stressed) mean of the variable $X^{j_0}$ while the former (unstressed) mean is $t_0$. In this case,
plotting $\mathrm{ER}_{t}$ as a function of $t-t_0$ highlights the variables which produce the largest amount of confusion in the error, \textit{i.e.} those for which small or large values provide the most variability among the two predicted classes, hampering the prediction error rate. 
The False and True Positive Rates may alternatively be represented using
\[
\mathrm{FPR}_{t}
=
\frac{
	\frac{1}{n}
	\sum_{i=1}^n
	\lambda^{(t)}_i
	\mathbbm{1}_{\left\{  Y_{i} \neq 1 \right\}}
	\mathbbm{1}_{\left\{  f(X_i) = 1 \right\}}
}
{
	\frac{1}{n}
	\sum_{i=1}^n
	\lambda^{(t)}_i
	\mathbbm{1}_{\left\{  f(X_i) = 1 \right\}}
}
\]
and
\[
\mathrm{TPR}_{t}
=
\frac{
	\frac{1}{n}
	\sum_{i=1}^n
	\lambda^{(t)}_i
	\mathbbm{1}_{\left\{  f(X_{i}) = 1 \right\}}
	\mathbbm{1}_{\left\{  Y_{i} = 1 \right\}}
}
{
	\frac{1}{n}
	\sum_{i=1}^n
	\lambda^{(t)}_i
	\mathbbm{1}_{\left\{  Y_{i} = 1 \right\}}
}.
\]
Again with $\Phi$ as in Theorem \ref{th:mainenpractice},
a ROC curve corresponding to perturbations of the variable $j_0$ can then be obtained by plotting pairs $(\mathrm{FPR}_{t},\mathrm{TPR}_{t})$ for a large number of $t \in \mathbb{R}$. We then obtain the evolution of both errors when $t$ evolves, which allows a sharper analysis of the error evolution (see \textit{e.g.} Section~\ref{ssec:ROCcurves}).
Finally, the variables influence  on the predictions may be quantified by computing the portion of predicted $1$s
\[
\mathrm{P1}_{t}
=
\frac{1}{n}
\sum_{i=1}^n
\lambda^{(t)}_i
f(X_i)
\]
which we suggest to plot similarly as $\mathrm{ER}_{t}$, with $\Phi$ as in Theorem \ref{th:mainenpractice} (see Figure~\ref{fig:ResultsAdultIncome}-(top)).
The figures representing $\mathrm{P1}_{t}$ make it possible to simply understand the particular influence of the variables to obtain a  decision $\hat{Y}=1$, whatever the veracity of the prediction. Importantly, they point out which variables should be positively or negatively modified in order to change a given decision.

\subsection{The case of multi-class classification}\label{ss:multiclassclassif}
We now consider the case of a classification into $q$ different categories, \textit{i.e.} where  $Y_i$ and $f(X_{i})$ belong to $  \{ 1 , \ldots , q\}$ for all $i = 1, \ldots,n$, and where $q \in \mathbb{N}$ is fixed.

In this case, the strategy described for binary classification can naturally be extended using
\[
\mathrm{Pj}_{t}
=
\frac{1}{n}
\sum_{i=1}^n
\lambda^{(t)}_i
\mathbbm{1}_{\left\{ f(X_i) = j \right\}},
\]
which denotes the portion of individuals assigned to the class $j$.
%

\subsection{Using quantiles to compare multiple mean changes}\label{ss:multmeanchanges}

Consider the case where $\Phi$ is given as in Theorem \ref{th:mainenpractice}, with $\Phi(X^1,\ldots,X^p,\hat{Y},Y) = X^{j_0}$, and where we want to plot the quantities of interest of Sections \ref{ss:binary:classif} and \ref{ss:multiclassclassif}, as a function of $t - t_0$, for all the values of $j_0 =1 , \ldots,p$. In this case, an issue is that the interpretation of $t-t_0$ depends on the order of magnitude of the variable $j_0$, and thus changes with $j_0$.

In order to compare values of $t-t_0$ across different variables, we suggest a common parametrization of $t-t_0$ for $j_0 = 1,\ldots,p$. We consider the empirical quantile function $q_{j_0}$  associated to the variable $X^{j_0}$, so
$q_{j_0}(\rho)= X^{j_0}_{\sigma([n\rho])}$ for $0 \leq \rho < 1$ and where $\sigma(.)$ is a function ordering the sample, \textit{i.e.} $X_{\sigma(0)}^{j_0} \leq X_{\sigma(1)}^{j_0} \leq \ldots \leq X_{\sigma(n-1)}^{j_0}$. Then, the range of the stressed mean $t$ will be in $[q_{j_0}(\alpha),q_{j_0}(1-\alpha)]$, where $\alpha \in (0,1/2)$ (a typical value is $0.05$).

We then tune $t - t_0$ as equal to $\epsilon_{j_0,\tau}$, where $\epsilon_{j_0,\tau}=\tau (  t_0 - q_{j_0}(\alpha))$ if $\tau \in [-1,0]$, and $\epsilon_{j_0,\tau}= \tau (  q_{j_0}(1-\alpha) - t_0  )$ if $\tau  \in [0,1]$. Parameter $\tau$ therefore allows to intuitively parametrize the level of stress whatever the distribution of the
$ \{ X_{1}^{j_0} , \ldots , X_{n}^{j_0} \}$. More precisely, $\tau = 0$ yields no change of mean, $\tau = -1$ changes the mean from $t_0$ to the (small) quantile $q_{j_0}(\alpha)$ and $\tau = 1$ changes the mean from $t_0$ to the (large) quantile $q_{j_0}(1-\alpha)$.

For $j_0 =1 , \ldots,p$ and $\tau \in [-1,1]$, we thus naturally suggest to compute

\begin{equation} \label{eq:Qjzerotau}
	\mathrm{ER}_{j_0,\tau},
	~
	\mathrm{FPR}_{j_0,\tau},
	~
	\mathrm{TPR}_{j_0,\tau},
	~
	\mathrm{P1}_{j_0,\tau},
	~
	\mathrm{Pj}_{j_0,\tau}
\end{equation}
that are defined as
$\mathrm{ER}_{t},
\mathrm{FPR}_{t},
\mathrm{TPR}_{t},
\mathrm{P1}_{t},
\mathrm{Pj}_{t}
$ in Sections \ref{ss:binary:classif} and \ref{ss:multiclassclassif}, with $\Phi(X^1,\ldots,X^p,\hat{Y},Y) = X^{j_0}$ and $t = t_0 + \epsilon_{j_0,\tau}$ as explained above. For a given $\tau$, it makes sense to compare the indicators in \eqref{eq:Qjzerotau} across $j_0 = 1,\ldots,p$.
For instance, one can plot $\mathrm{ER}_{j_0,\tau}$ as a function of $\tau$ for $\tau \in [-1,1]$ and for each $j_0 \in \{ 1 , \ldots , p\}$, as shown in Figure~\ref{fig:ResultsAdultIncome}-(bottom). Remark that $\tau=0$ corresponds to the algorithm performance baseline, without any perturbation of the test sample.

\subsection{\textcolor{black}{The case of regression}}
\label{ssec:regressionCase}

We consider now the case of a real valued regression where $Y_i,f(X_{i}) \in \mathbb{R} $ for $i = 1 \ldots,n$. In order to understand the effects of each variable, first we consider the mean criterion
\[
\mathrm{M}_{j_0,\tau}
=
\frac{1}{n}
\sum_{i=1}^n
\lambda^{(j_0,\tau)}_i
f(X_i),
\] which will indicate how a change in the variable will modify the output of the learned regression ($\tau$ is explained above and we let $\lambda^{(j_0,\tau)}_i$, $i=1,\ldots,n$, be the corresponding weights). Second
the variance criterion
\[
\mathrm{V}_{j_0,\tau}
=
\frac{1}{n}
\sum_{i=1}^n
\lambda^{(j_0,\tau)}_i
\left(
f(X_i)
-
\mathrm{M}_{j_0,\tau}
\right)^2
\] is meant to study the stability of the regression with respect to the perturbation of the variables. Finally
the root mean square error (RMSE) criterion
\[
\mathrm{RMSE}_{j_0,\tau}
=
\sqrt{
	\frac{1}{n}
	\sum_{i=1}^n
	\lambda^{(j_0,\tau)}_i
	\left(
	f(X_i)
	-
	Y_{i}
	\right)^2
}
\] is analogous to the classification error criterion since it enables to detect possibly misleading or confusing variables when learning the regression. \\
For each $j_0 \in \{ 1 , \ldots , p\}$, these three criteria can be plotted as a function of $\tau$ for $\tau \in [-1,1]$.

\section{Results}
\label{res:results}

In this section, we illustrate the use of the indices obtained using the entropic projection of samples on \textcolor{black}{multiple application} cases. In Section~\ref{res:resultsTwoClass}, the \textit{Adult Income} dataset\footnote{https://archive.ics.uci.edu/ml/datasets/adult} is considered, where $X$ represents $n=32000$ observations of dimension $p=14$ and $Y$ has $2$ classes. The results of Section~\ref{res:classifImages} are first obtained on the \textit{MNIST} dataset\footnote{http://yann.lecun.com/exdb/mnist/}, where $X$ represents $n=60000$ gray level images of  $p=784$ pixels and $Y$ has $10$ classes, and then obtained on the
\textit{Large-scale CelebFaces Attributes (CelebA)} Dataset\footnote{http://mmlab.ie.cuhk.edu.hk/projects/CelebA.html} where
$X$ represents $n=200000$ RGB images of  $4096$ pixels, so $p=4096*3=12288$, and $Y$ has $2$ classes.
\textcolor{black}{Section~\ref{sssec:SimulatedData} addresses synthetic data and shows that our method yields the same hierarchy of variables as that given by the (unknown) true data generating process. Also for synthetic data,
Section~\ref{sssec:SimulatedData:corr:caus} shows that our method enables to discriminate between correlation and causation.
Section \ref{ssec:ROCcurves} provides an analysis of ROC curves in binary classification. Section~\ref{ssec:resultsIRIS} studies the effect of 4 variables on the classification of  the well known \textit{Iris} dataset.
Section \ref{ss:regressionCase:appli} addresses the \textit{Boston Housing} dataset in regression. Finally, the computational burden is evaluated in Section~\ref{ssec:ComputaBurden}.}

Importantly, the Python code to reproduce these experiments is freely available\footnote{\url{https://gems-ai.aniti.fr/}}.

\subsection{\textcolor{black}{Binary} classification: {\it Adult Income} dataset}\label{res:resultsTwoClass}

\begin{figure*}[t]
	\begin{center}
		\includegraphics[width=0.90\linewidth]{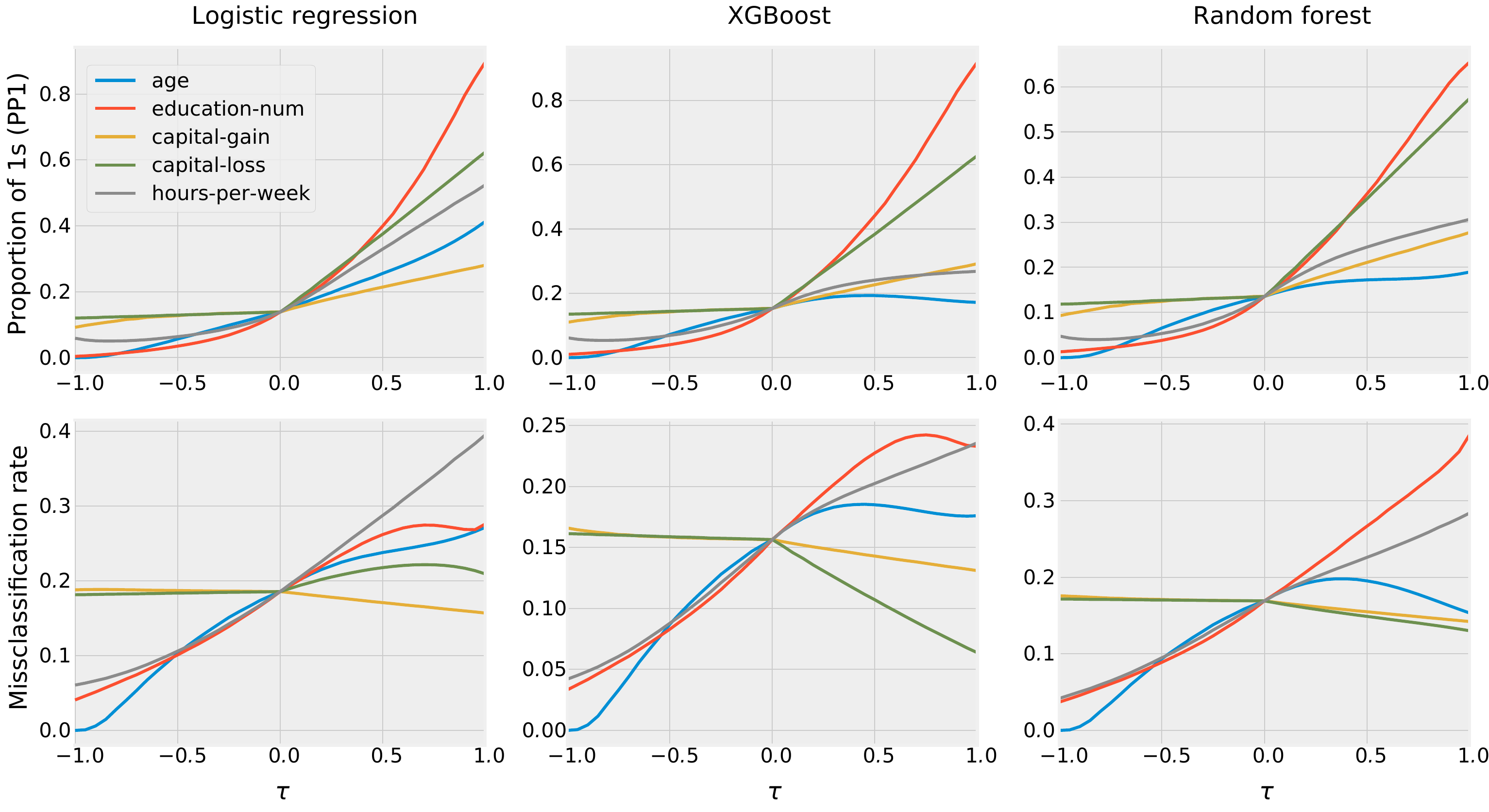}
		\caption{
			Results of Section~\ref{res:resultsTwoClass} on the \textit{Adult income} dataset.
			{\bf (Top - PP1s)} Portion of predicted ones (\textit{i.e.} High Incomes) with respect to the explanatory variable perturbation  $\tau$.
			{\bf (Bottom - Mis. Rate)} Classification errors with respect to $\tau$.
			There is no perturbation if $\tau = 0$. The larger (resp. the lower) $\tau$, the larger (resp. the lower) the values of the selected explanatory variable.
		}
		\label{fig:ResultsAdultIncome}
	\end{center}
\end{figure*}

\begin{figure*}[t]
	\begin{center}
		\includegraphics[width=0.70\linewidth]{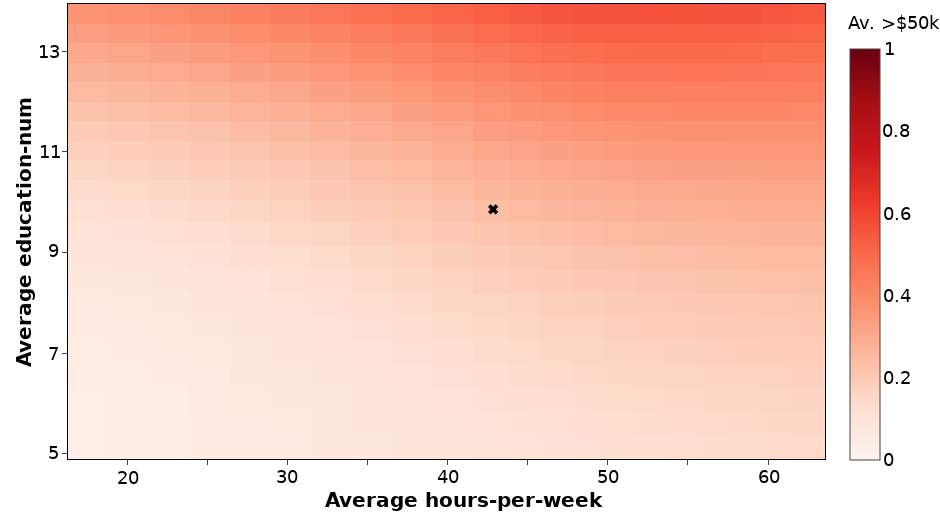}
		\caption{
			Result of Section~\ref{res:resultsTwoClass} on the \textit{Adult Income} dataset with the \textit{XGboost} classifier.
			The portion of predicted ones (\textit{i.e.} High Incomes) with respect to the variables \textit{education-num} and \textit{hours-per-week} simultaneously is represented. The black cross represents the average value in the unstressed dataset.
		}
		\label{fig:ResultsAdultIncome_two_conditions}
	\end{center}
\end{figure*}

In order to illustrate the performance of our procedure, we first consider the \textit{Adult Income} dataset. It is made of $n=32000$ observations represented by $p=14$ attributes ($6$ numeric and $8$ categorical), each of them describing an adult.
We specifically interpret the influence of $5$ numeric variables on the categorical output variable representing whether each adult has an income higher ($Y_i=1$) or lower ($Y_i=0$) than $50000\$$ per year.

We first trained three different classifiers (Logit Regression, XGboost and Random Forest\footnote{\textit{R} command \textit{glm} and packages \textit{xgboost} and \textit{ranger}.}) on $25600$ randomly chosen observations ($80\%$ of the whole dataset). We then  performed the proposed entropic projection strategy for each learned classifier on a test set made of the $6400$ remaining observations.
Detailed results are shown in Figure~\ref{fig:ResultsAdultIncome}.
Instead of only quantifying a score for each variable, we display in this figure the evolution of the algorithm confronted at gradually lower or higher values of $\tau$ (see Section \ref{ss:multmeanchanges}) for each variable.  The curves were computed using $21$ regularly sampled values of $\tau$ between $-1$ and $1$ with  $\alpha=0.05$. For each variable, the weighted observations were therefore stressed so that their mean is contained between the $0.05$ and $0.95$ quantiles of the original (non-weighted) values distribution in the test set.
Note that for a quick and quantified overview of the variables response to a positive (resp. negative) stress, the user can simply interpret the difference of the response for $\tau=1.$ and $\tau=0$ (resp. $\tau=0$ and $\tau=-1.$), as illustrated
in Section~\ref{res:classifImages} in the image case.

\paragraph{Influence of the variables on the decision rule}
We present in Figure~\ref{fig:ResultsAdultIncome}~\textit{(Top)} the role played by each variable in the portion of predicted ones (\textit{i.e.} high incomes) for the \textit{Logit Regression},  \textit{XGboost} and  \textit{Random Forest} classifiers.
The curves in Figure~\ref{fig:ResultsAdultIncome}~\textit{(Top)}  highlight the role played by the variable \textit{education-num}. The more educated the adult, the higher his/her income will be. The two variables \textit{capital-gain} and \textit{capital-loss} are also testimonial of high incomes since the adults having large incomes can obviously have more money than others on their bank accounts, or may easily contract debts, although the contrary is not true. It is worth pointing out the role played by the age variable which appears clearly in the figure: young adults earn increasingly large incomes with time, which is well captured by the decision rules (left part of the red curves).

We emphasize that these curves  enable to intuitively interpret the complex trends captured by \textit{black-box} decision rules.
They indeed  quantify non-linear effects of the variables and very different behaviors depending on whether the variables increase or decrease.

Contrary to methods that study the influence of the variables by computing information theory  criteria between different outputs of the algorithm for changes in the variables (see in \citet{skater}), the  variable changes we use are plausible since the stressed distributions are as close as possible to the initial distribution. Finally, we work directly on the real black-box model and do not approximate it by any surrogate model, as in \citet{ribeiro2016should}.

It can finally be useful to explain the role played by two variables by stressing them simultaneously. We show for instance in Figure~\ref{fig:ResultsAdultIncome_two_conditions} the role played by the variables \textit{education-num} and \textit{hours-per-week} with the \textit{XGboost} classifier. This makes clear the fact that the variable \textit{education-num} has more influence than \textit{hours-per-week} to predict high incomes, as already shown in Figure~\ref{fig:ResultsAdultIncome}~\textit{(Top)}. This additionally shed lights on the fact that the persons with the highest \textit{education-num} values will have slighly decaying predictions if they work more than 52 hours per week, which is not the case for the persons with less than 12 years of academic education.

\paragraph{Influence of the variables on the accuracy of the classifier}
Besides the influence of the variables on the algorithm outcome, it is worth studying their influence on the accuracy or veracity of the model.
We then present in Figure~\ref{fig:ResultsAdultIncome}~\textit{(Bottom)} the evolution of the classification error when each variable is stressed by $\tau$.  The three sub-figures (one for each prediction model) represent the evolution of the error confronted to the same modification of each variables. The error of the method on the original data is obtained for $\tau=0$.  Such curves point out which variables have the strongest influence on prediction errors. Such result may be used to temper the trust in the forecast depending on the values of the variables.

As previously, the curves appear as  more informative than single scores:
the three models enable to select  the same couple of variables that are important for the accuracy of the prediction when they increase \textit{i.e.} education number and numbers of hours worked per week.  The latter makes the prediction task the most difficult when it is increased.
Indeed, the persons working a large number of hours per week may not always increase their income, since it relies on different factors. People with high income however usually work a large number of weekly hours. Hence, these two variables play an important role in the prediction and their changes impact the prediction error.
In the same flavor,  more insight on the error terms could be obtained by dealing with  the evolution of the False Positive Rate and True Positive Rate as presented in Section~\ref{ssec:ROCcurves}.

\subsection{Image classification}\label{res:classifImages}

\subsubsection{\textit{MNIST} dataset}

We now  measure the influence of pixel intensities in image recognition tasks. Each pixel intensity is treated as a variable and the stress is used to saturate the intensities towards one side of their spectrum (red if $\tau = 1$ or blue if $\tau = -1$).

\begin{figure*}[htb]
	\begin{center}
		\includegraphics[width=0.90\linewidth]{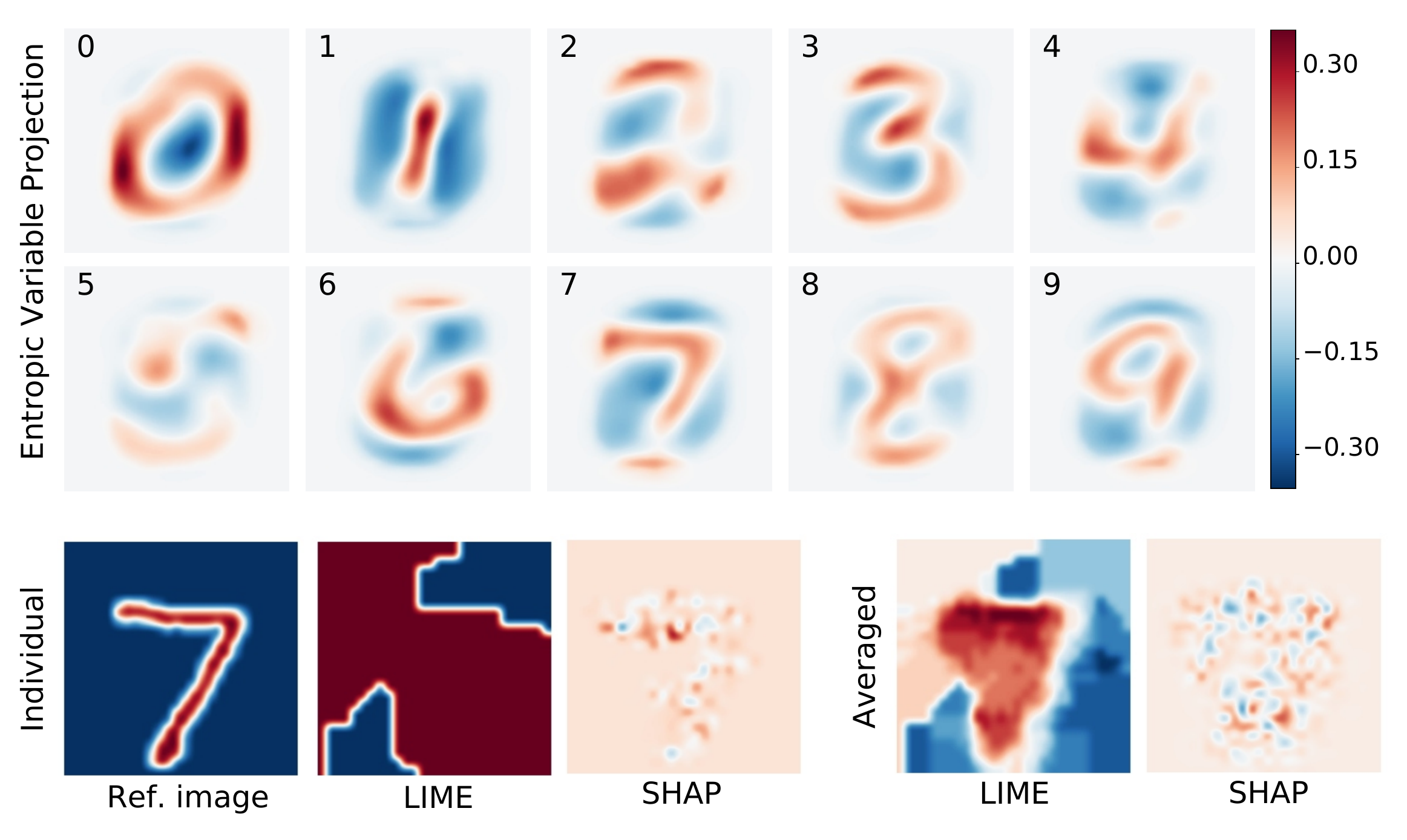}
		\caption{\textbf{(top)} Pixel contributions towards each digit according to our entropic projection method.
			\textbf{(bottom-left)} Pixel contributions to predict seven in an individual image representing a seven, using the LIME and SHAP packages \textcolor{black}{of local explainability}.
			\textbf{(bottom-right)} Average pixel contributions to predict seven in all images of the \textit{MNIST} test set representing a seven, using the LIME and SHAP packages \textcolor{black}{(that is, aggregating the local explanations of LIME and SHAP into global ones)}.}
		\label{fig:ResultsMNISTDiff}
	\end{center}
\end{figure*}

We specifically trained a CNN (Convolutional Neuron Network) on the \textit{MNIST} dataset using a typical architecture that can be found on the Keras documentation\footnote{https://keras.io/examples/mnist\_cnn}. The CNN was trained on a set of 60000 images whilst the predictions were made on another 10000 images. It achieved a test set accuracy north of 99\%. For each of the 784 pixels $j_0$, we computed the perturbed distributions in the cases where $\tau$ equals $-1$ as well as $1$, using the method of Section~\ref{ss:multmeanchanges}. The prediction portions of each of the 10 digits was then computed using the method of  Section~\ref{ss:multiclassclassif}.
The whole process took around 9 seconds to run on a modern laptop (Intel Core i7-8550U CPU @ 1.80GHz, 24GB RAM) running Linux. The results are presented in Figure~\ref{fig:ResultsMNISTDiff}-(top).

The color of each pixel in Figure~\ref{fig:ResultsMNISTDiff}-(top) represents its contribution towards the prediction of each digit. For example, a value of 0.15 means that the CNN predicts on average this digit 15\% more often when the associated pixel is activated ($\tau = 1$) instead of having the background intensity ($\tau = -1$). Although our method is pixel-based, it is still able to uncover regions which the CNN uses to predict each digit. Likewise, redder regions contain pixels that are positively correlated with each digit. Note that the edges of each image don't change color because the corresponding pixels have no impact whatsoever on the predictions. The left part of number 5 has pixels in common with number 6. However, we are able to see that the CNN identifies 6s by using the bottom part of the 6, more so than the top stroke which it uses to recognize 5s. Likewise, according to the CNN, the most distinguishing part of number 9 is the part that links the top ring with the bottom stroke.

\begin{remark}[\textcolor{black}{Comparison with LIME and SHAP}]
\textcolor{black}{Recall that our explainability framework is global, as discussed in the introduction. As also discussed in the introduction, the two most popular explainability methods are} LIME \citep{ribeiro2016should}\footnote{https://github.com/marcotcr/lime} and
SHAP \citep{shap}\footnote{https://github.com/slundberg/shap}.
\textcolor{black}{These methods are local, meaning that they can provide different explainability plots for each different digit image of the \textit{MNIST} dataset (while we provide the same explainability plots for all the digit images). Nevertheless, the popularity of these two methods justify comparing their interpretations with ours.}
 As an illustration, Figure~\ref{fig:ResultsMNISTDiff}-(bottom-left) represents the most influential pixels found with LIME and SHAP to predict a seven in an \textcolor{black}{individual} image of the \textit{MNIST} test set representing a seven. The results are not straightforward to interpret.
\textcolor{black}{To construct global explanations from the individual explanations of LIME and SHAP,} one can represent the average results obtained by using LIME or SHAP over all  images of the \textit{MNIST} test set representing a seven, as represented in Figure~\ref{fig:ResultsMNISTDiff}-(bottom-right). Note that the computations required take about 7 and 10 hours using LIME and SHAP, respectively, which is much longer than when using our method (9 seconds). Compared to our method, averaged results are also less resolved for LIME and harder to interpret for SHAP. Our method can also natively compute  other properties of the black-box decision rules, as described in Section~\ref{s:main2}.
\end{remark}

\subsubsection{\textit{CelebA} dataset}

We now present further results obtained on the \textit{Large-scale CelebFaces Attributes (CelebA)} Dataset \citep{liu2015faceattributes}. It contains more than 200000 celebrity RGB images, each with 40 binary attribute annotations, such as  \textit{Eyeglasses},  \textit{PaleSkin},  \textit{Smiling}, \textit{Young},  \textit{Male} or \textit{Attractive}. It is important to mention here that the images of the \textit{CelebA} dataset cover large pose variations and background clutter. This makes this dataset far more complex to analyze than the \textit{MNIST} dataset.
In order to explain the decision rules of a state-of-the-art neural-network architecture, we trained a ResNet-18 convolutional Neural Network  \citep{He2016DeepRL} to predict the attribute \textit{Attractive} based on the \textit{CelebA} images. The portion of good predictions on the training set was 0.95 and 0.92 for females and males, respectively. This portion was also 0.86 and 0.78 on the test set  for females and males, which is good for such a subjective attribute. This suggests that the persons who labeled the data were relatively coherent when choosing who was attractive or not.

\begin{figure*}[htb]
	\begin{center}
		\includegraphics[width=0.70\linewidth]{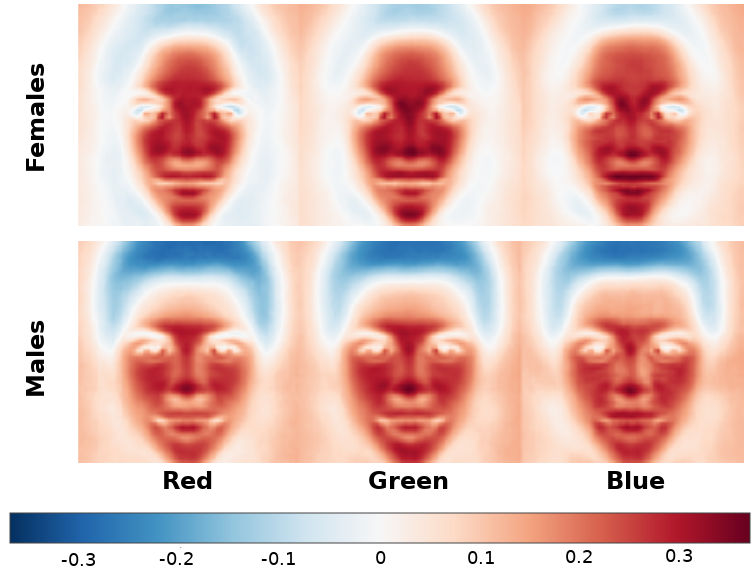}
		\caption{Pixel contributions to explain the decision rules of a ResNet-18 neural network trained to predict wether someone is considered as attractive or not. The flexibility of the decision rules in different contexts is emphasized by using the entropic projection strategy on images representing females and males separately.}
		\label{fig:ResultsCelebA}
	\end{center}
\end{figure*}

We then used almost the same strategy as on the \textit{MNIST} dataset to explain the decision rules learned by the ResNet-18 neural-network. In order to show the flexibility of its decision rules in two different contexts, we however used the entropic projection strategy on the images representing females and males separately. The results are presented in Figure~\ref{fig:ResultsCelebA}. Note that they were obtained by using all the 200000 images of the dataset and required about three minutes of computations.
As shown in Figure~\ref{fig:ResultsCelebA}, it is clear that the trained  neural-network takes more into account the hairs for males than for females to make its predictions. It also uses more the contour of the face for females than for males (in particular for the pixels corresponding to the cheeks).

\subsection{\textcolor{black}{Binary classification: simulated data}}
\label{ssec:binary:simulated}

\subsubsection{\textcolor{black}{Recovering hierarchy of variables}} \label{sssec:SimulatedData}

\begin{figure*}
	\begin{center}
		\includegraphics[width=0.40\linewidth]{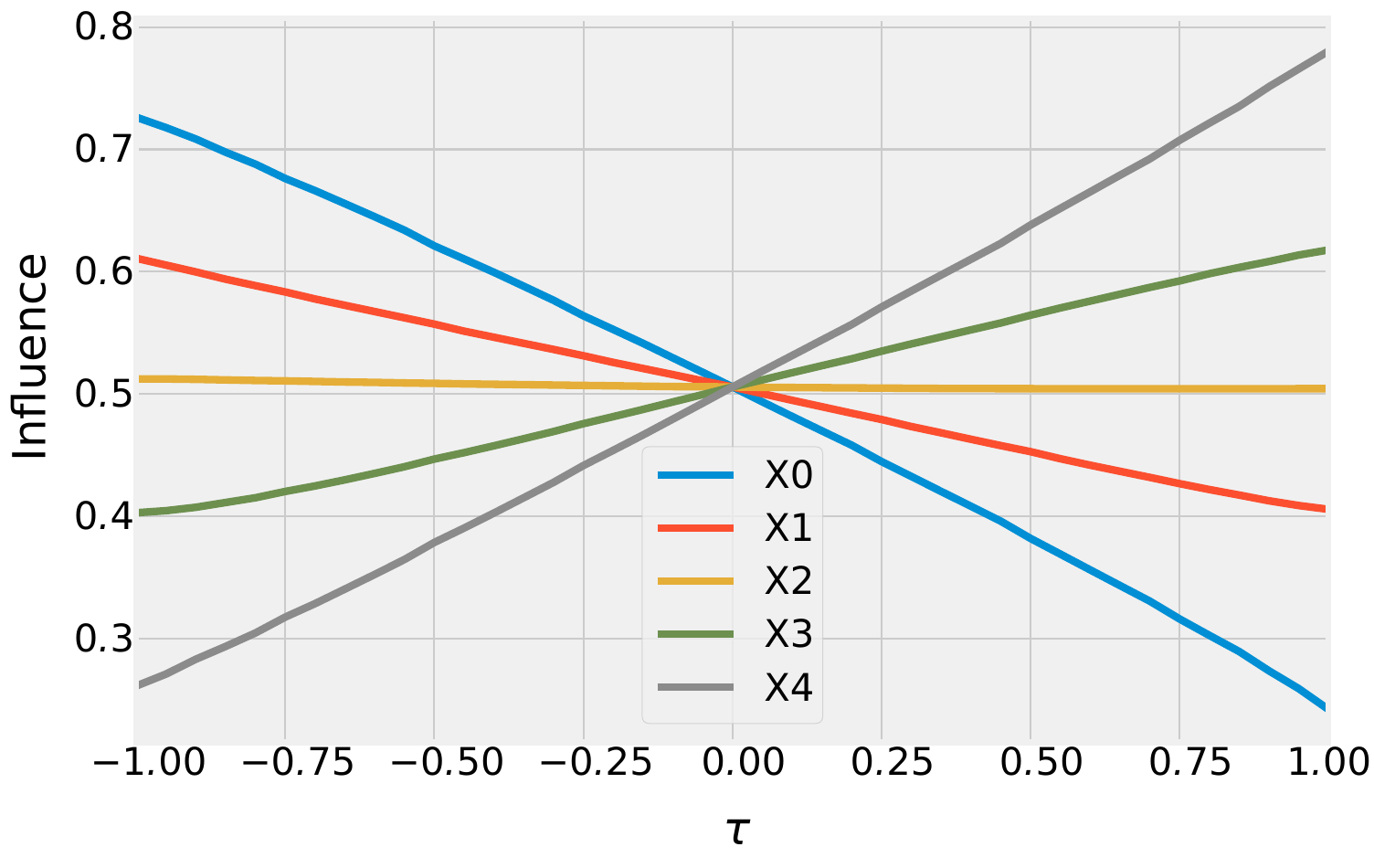}
		\caption{
			Portion of ones found on synthetic data generated using a logistic regression model.}
		\label{fig:ResultsSynData}
	\end{center}
\end{figure*}

In order to further show that our procedure is able to properly recover the characteristics of machine learning algorithms,  we again tested it on synthetic data. We have run an experiment with $p=5$ variables and $n=10^6$ observations, where synthetic data are generated using a logistic regression model, with independent regressors and coefficient vector equal to $\beta = (-4,-2,0,2,4)$. \textcolor{black}{That is, for $i=1,\ldots,n$,  $X_i \sim \mathcal{N}(0 , I_p)$ and $\mathbb{P}(Y_i = 1 | X_i) = \psi(\langle \beta , X_i \rangle)$ with $\psi(t) = 
e^{t} / (1+e^{t})$. There is independence between data points $i = 1 , \ldots , n$. We consider the portion of ones indicator of Section \ref{ss:binary:classif} with $f(X_i)$ replaced by $Y_i$, and stress the means of the input variables as in Theorem \ref{th:mainenpractice}.}
Figure~\ref{fig:ResultsSynData} clearly shows that our method enables to recover the signs and the hierarchy of the coefficients.

\subsubsection{\textcolor{black}{Discriminating between correlation and causation}} \label{sssec:SimulatedData:corr:caus}

\begin{figure*}
	\begin{center}
		\includegraphics[width=0.40\linewidth,height=5cm]{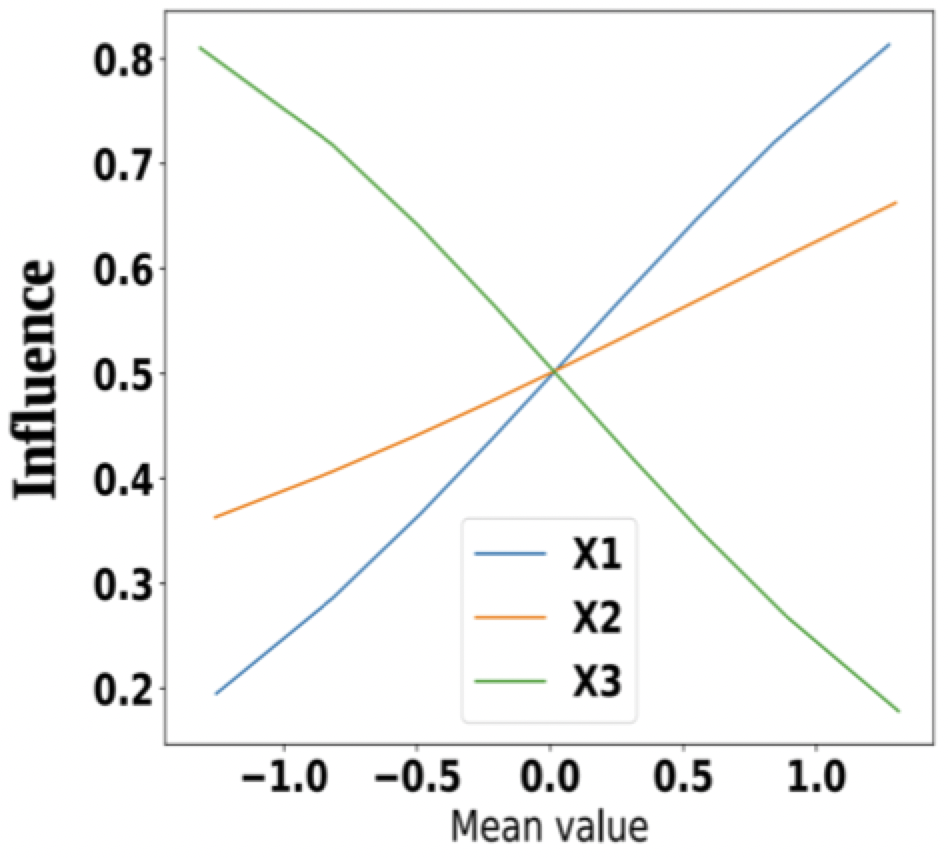}
		\includegraphics[width=0.40\linewidth,height=5.05cm]{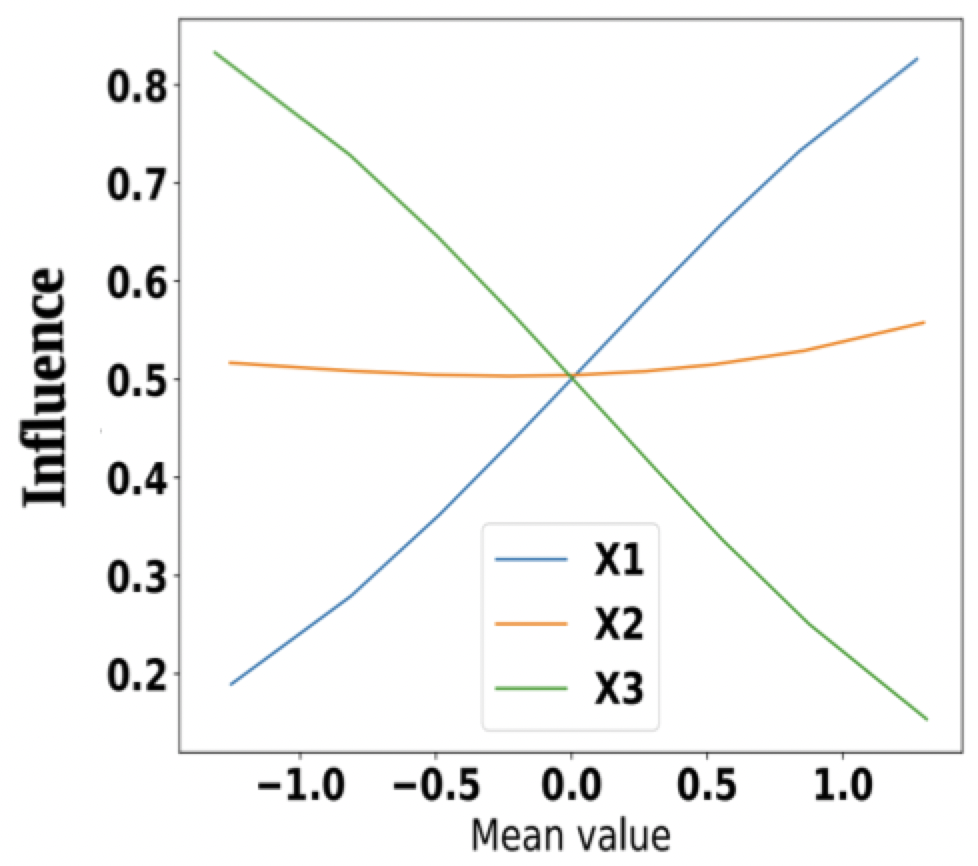}
		\caption{
		\textcolor{black}{
			Portion of ones found on synthetic data generated using a binary regression model with correlated variables. Left: correlation is detected for the second variable. Right: absence of causality is detected for the second variable.} 
		}
		\label{fig:ResultsSynData:causality}
	\end{center}
\end{figure*}

\textcolor{black}{
We run a similar experiment as before, but now with $p=3$ variables and $n=2000$ observations. 
The data $(X_i,Y_i)_{i=1,\ldots,n}$ are sampled as before with the only difference that $X_i$ is sampled from the 
\[
\mathcal{N} 
\left( 
\begin{pmatrix} 
0 \\ 0 \\ 0
\end{pmatrix},
\begin{pmatrix} 
1 & 0.5 & 0  \\
0.5 & 1 & 0 \\
0 & 0 & 1
\end{pmatrix}
\right)
\]
distribution, that we have $\beta = (1,0,-1)$ and that we have $\Psi(t) = \Phi(10t)$ with $\Phi$ the standard Gaussian c.d.f. Hence, the second variable is important in the sense that it is positively correlated with the first one, that itself has a positive regression coefficient. This is correctly detected by stressing the means of the input variables as in Theorem \ref{th:mainenpractice} and plotting the portion of ones as before, on the left-hand side of Figure \ref{fig:ResultsSynData:causality}.}

\textcolor{black}{However, the second variable has no causal influence. That is, if variables one and three are fixed and variable two is changed, the conditional distribution of the output is unchanged. The right-hand side of Figure \ref{fig:ResultsSynData:causality} shows that our framework can detect this absence of causation. There, we plot again the portion of ones. For the line plot of the first variable, we stress its mean, we impose an unchanged mean for the second one, and we impose a zero covariance between these two variables, using Corollary \ref{cor:stress:covariance}. For the second line plot, we do the same but the mean of the first variable is unchanged and the mean of the second one is stressed. For the third line plot, the means of the two first variables are unchanged, their covariance is imposed to zero, and the mean of the third variable is stressed. We see that the first variable remains influential (it indeed has a causal influence on the output) but the second variable becomes non-influential.\vskip .1in  Hence, our framework has enabled to answer the question {\it what if two variables were uncorrelated} to show that one variable is correlation-influential but not causality-influential. To the best of our knowledge, this possibility of discriminating correlation and causality, in a model-agnostic way and for any machine learning predictor, does not appear elsewhere in the literature. }

\textcolor{black}{For comparison with another model-agnostic explainability tool, we now consider the Shapley values of the three variables \citep{owen_shapley_2017,shapley_value_1953,shap}.
To numerically evaluate the Shapley values based on the same dataset as before we use the \textit{shapleySubsetMc} function from the \textit{R} package \textit{sensitivity}, see also \cite{broto2020variance}. The computation time is approximately 110 seconds. In contrast, with our suggested method, this time was approximately 4 seconds. We thus see here a computational benefit of our suggested approach. The computed Shapley values of the three variables are approximately $0.45, 0.06, 0.49$. Hence, we see that variables one and three are close to equally important and variable two has a much smaller non-negligible importance. We have a single importance indicator for variable two which does not allow to discriminate between correlation and causation, contrary to our suggested framework.
}

\subsection{Other indices \textcolor{black}{for binary classification}: ROC Curves}\label{ssec:ROCcurves}

\begin{figure*}
	\begin{center}
		\includegraphics[width=0.99\linewidth]{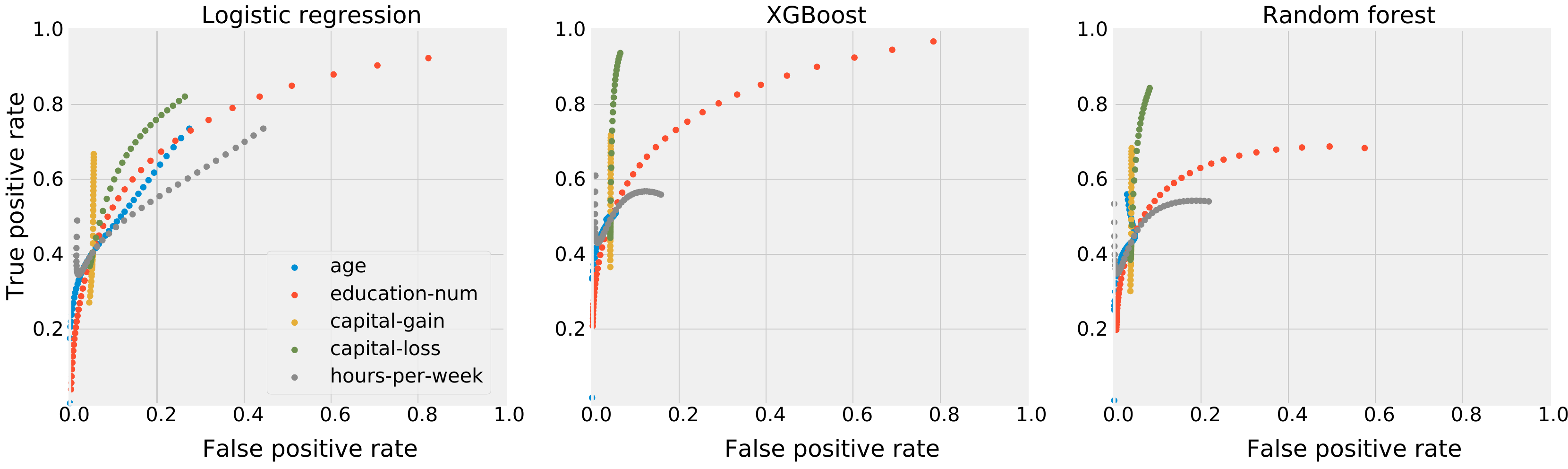}
		\caption{
			Evolution of ROC curves  in the \textit{Adult Income} dataset (Section \ref{res:resultsTwoClass}). As for the classification errors, we observe that large values of the variable \textit{hoursWeek} make the classification difficult.
		}
		\label{fig:ResultsAdultIncomeROC}
	\end{center}
\end{figure*}

In the case of two class classification on the \textit{Adult Income} dataset (Section \ref{res:resultsTwoClass}), we have shown the evolution of the classification error when the stress parameter $\tau$ increases. Such results can straightforwardly be extended to True and False Positive Rates, which are commonly represented in ROC curves, that we display in Figure \ref{fig:ResultsAdultIncomeROC}. Each point of these curves corresponds to the False Positive Rate and the True Positive Rate, for a sample corresponding to each $\tau$ and each variable. All curves cross at the same point which corresponds to $\tau=0$. It therefore becomes possible to study the evolution of each criterion.

\subsection{\textcolor{black}{Multi-class classification: {\it Iris} dataset}} \label{ssec:resultsIRIS}

\begin{figure*}
	\begin{center}
		\includegraphics[width=0.99\linewidth]{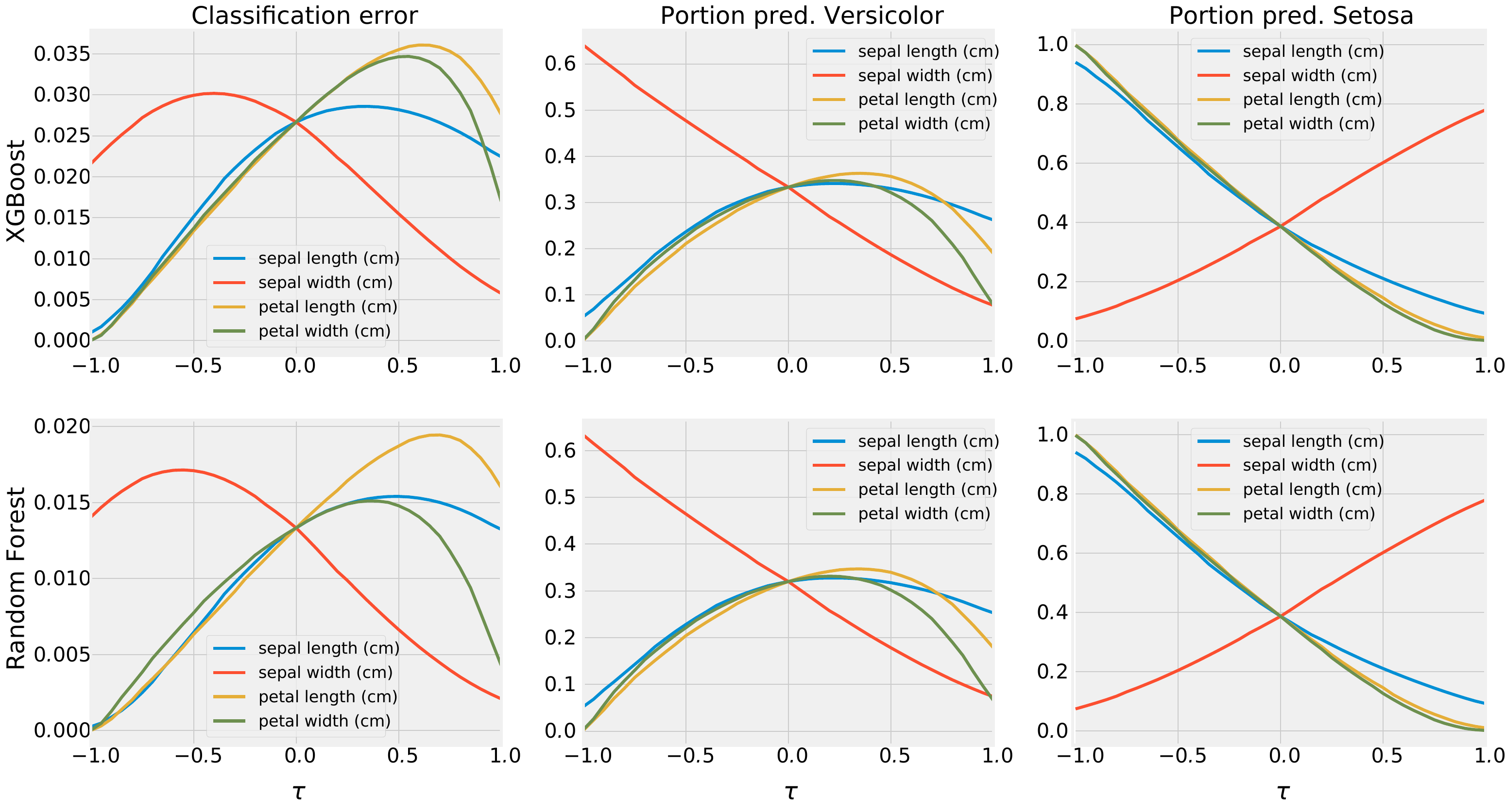}
		\caption{
			Evaluation of the classification error and the prediction with respect to the explanatory variable perturbation  $\tau$, on the \textit{Iris} dataset. The quantity $\tau$ and the lines have the same signification as in Figure \ref{fig:ResultsAdultIncome}.
			{\bf (Top)}
			XGBoost Model. The sepal width enables to differenciate the \textit{Setosa} class.
			{\bf (Bottom)}
			Random Forest Model. The sepal width again enables to differenciate the \textit{Setosa} class.
		}
		\label{fig:ResultsIRIS}
	\end{center}
\end{figure*}

As an additional assesment of the method on very well known and simple data, we now consider the \textit{Iris} dataset\footnote{https://archive.ics.uci.edu/ml/datasets/iris}. This dataset is composed of 150 observations with 4 variables used to predict a label into three categories: \textit{setosa}, \textit{versicolor}, \textit{virginica}. To predict the labels, we used an \textit{Extreme Gradient Boosting} model and a \textit{Random Forest} classifier. Results are show in Figure~\ref{fig:ResultsIRIS}. We present for both models  the classification error and portion of class prediction, showing the effects of increasing or decreasing the 4 parameters, \textit{i.e.} the width or the length of the sepal or petal. As expected, we recover the well known result that the width of the sepal is the main parameter which enables to differentiate the class \textit{Setosa} while the differentiation between the two other remaining classes is less obvious.

\subsection{\textcolor{black}{Regression: {\it Boston Housing} dataset}} \label{ss:regressionCase:appli}

\begin{figure*}
	\begin{center}
		\includegraphics[width=0.90\linewidth]{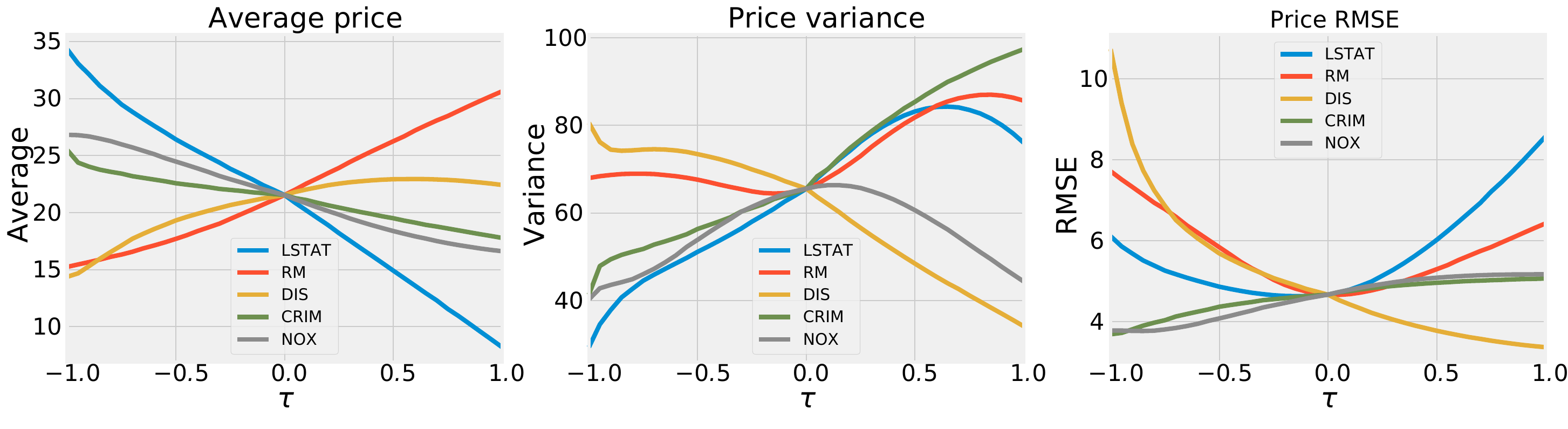}
		\caption{
			Results obtained on the \textit{Boston Housing} dataset with Random
			Forest. The explanatory variable perturbation  $\tau$  has the same signification as in Figure \ref{fig:ResultsAdultIncome}.
		}
		\label{fig:ResultsBoston}
	\end{center}
\end{figure*}

We use now our strategy on the \textit{Boston Housing} dataset\footnote{\url{https://www.kaggle.com/c/boston-housing}}. This dataset deals with house prices in Boston. It contains 506 observations with 13 variables that can be used to predict the  price of the house to be sold.  When considering an optimized Random Forest algorithm,  the importance calculated as described in \citet{breiman2001random} enables to select  the 5 most important variables  as follows:
\textit{lstat} (15227), \textit{rm} (14852), \textit{dis} (2413), \textit{crim} (2144) and \textit{nox} (2042).
Remark that the coefficients obtained using a linear model would lead to similar interpretations, with the the 5 most important variables  as follows:
\textit{lstat} (-3.74), \textit{dis} ( -3.10),   \textit{rm} (2.67),   \textit{rad} (2.66),   \textit{tax} (-2.07).

As shown in Figure~\ref{fig:ResultsBoston}, our analysis goes further than these scores.
In particular we point out the non linear influence of the variables depending on whether they are high or low. For instance the average number of rooms in a house  (variable \textit{rm}) is an important factor that makes the price increase in the case of large houses ($\tau>0.$ in Figure~\ref{fig:ResultsBoston}~\textit{(Average)}). Interestingly, this is far less the case for  smaller houses ($\tau<0.$ in Figure~\ref{fig:ResultsBoston}~\textit{(Average)}) since there are other arguments than the number of rooms to keep a high price in this case.

Note that when the number of variables is large,  the presence of too many curves may make the graph difficult to understand. In this case, scores that represent average individual evolutions on given ranges of  $\tau$ values for each variables can be computed. Then the highest and lowest scores correspond to the most influential variables on the predictions. For instance, we represent in Table~\ref{tab:simData} the evolution of the Average curves in Figure~\ref{fig:ResultsBoston} between $\tau=-0.5$ and $\tau=0$, as well as between $\tau=0$ and $\tau=0.5$, which makes clearly understandable which the most influential variables are.
It is important to remark that our methodology still allows that the learned decision rules won't be mainly influenced by the same variables depending on whether they increase ($\tau>0$) or decrease ($\tau<0$).  In Table~\ref{tab:simData}, the more influential variables are indeed \textit{rm}, \textit{lstat} and \textit{zn} in the positive direction, while in the negative direction, the variables are  \textit{lstat}, \textit{black} and \textit{pratio}. Note that such variables are also cited in studies that rely on the LIME \citep{ribeiro2016should} or SHAP \citep{shap} packages, but the curves we present are more informative and rely on the same distributional input. \\

\begin{table}
	\begin{center}
		\begin{tabular}{ccc}
			\begin{tabular}{|c|}
				\hline
				$Mean_{0}-Mean_{-0.5}$\\
				\hline
				black (4.1)\\
				rm (3.0)\\
				dis (1.7)\\
				$\ldots$\\
				indus (-3.2)\\
				ptratio (-3.8)\\
				lstat (-5.1)\\
				\hline
			\end{tabular}
			&  $\null$ \quad $\null$&
			\begin{tabular}{|c|}
				\hline
				$Mean_{0.5}-Mean_{0}$\\
				\hline
				rm (6.80)\\
				zn (4.60)\\
				chas (2.74)\\
				$\ldots$\\
				indus (-3.05)\\
				tax (-3.18)\\
				lstat (-5.26)\\
				\hline
			\end{tabular}
		\end{tabular}
		\caption{\label{tab:simData} Most responsive variables to a positive or negative stress $\tau$ when estimating house prices. Scores are shown between  brackets and computed as the difference of the \textit{Mean} curves of Figure~\ref{fig:ResultsBoston} for \textbf{(left)} $\tau=-0.5$ and $\tau=0$, and  \textbf{(right)} $\tau=0$ and $\tau=0.5$.}
	\end{center}
\end{table}

\subsection{Evaluation of the computational burden}\label{ssec:ComputaBurden}

Recall  that our strategy only optimizes, for each of the $p$ variables, a function which evaluation cost is  $\mathcal{O}(n)$ with no additional outputs predictions out of the \textit{black box} machine learning algorithm. To quantify this, we show in Table~\ref{tab:simDataTime} the computational times dedicated to the analysis of synthetic datasets having a different amount of variables $p$ and observations $n$.
The variables interpretation was made using 21 values of $\tau$, leading to curves as \textit{e.g.} in Figure \ref{fig:ResultsAdultIncome}.
Computations were run with Python on a standard Intel Core i7 laptop with 24GB memory and no parallelization.
It appears that our strategy indeed has a $\mathcal{O}(np)$ cost, so  we then believe it may have a high impact to study the rules learned by black-box machine learning algorithms on large real-life datasets. Remark that when interpreting the influence of the pixel intensities on image test sets, as in Figure \ref{fig:ResultsMNISTDiff}, only 3 values of $\tau$ are used. The computations are therefore about 7 times faster. This is coherent with the 10 seconds required on 10000 \textit{MNIST} images of 28$\times$28 pixels in Section \ref{res:classifImages}. Note finally that a preliminary implementation of our method in R has lead to very similar results.

\begin{table}[ht]
	\begin{center}
		\begin{tabular}{|c|c||c|}
			\hline
			$p$&$n$&time (sec)\\
			\hline
			\hline
			10&10000&0.76\\
			\hline
			100&10000&7.79\\
			\hline
			1000&10000&82.5\\
			\hline
			10&100000&7.93\\
			\hline
			10&1000000&86.0\\
			\hline
		\end{tabular}
		\caption{\label{tab:simDataTime}Computational times required on synthetic datasets, where 21 levels of stress ($\tau$) were computed on each of the $p$  variables.}
	\end{center}
\end{table}

\section{Conclusion}\label{sec:discussion}
Explainability of \textit{black-box} decision rules in the machine learning paradigm  has many interpretations and has been tackled in a large variety of contributions. Here, we focused on the analysis of the variables importance and their impact on a decision rule, at the global scale of the entire dataset. We satisfy the important constraint, in the machine learning context, that the test input variables must follow the distribution of the learning sample, as closely as possible. Therefore, evaluating the decision rule at any possible point does not make any sense. To cope with this issue, we have  proposed an information theory procedure to bias the original variables without losing the information conveyed by the initial distribution. We proved that this solution amounts in re-weighting the observations of the test sample, leading to very fast computations and the construction of new indices to make clear the role played by each variable.

We highlight that our strategy can be seen as a \textit{what if} tool, as  counterfactual methods \citep{WachterHJLT18}. It indeed explains model decisions by quantifying how their outputs change when the machine learning data are transformed and thus captures some causal relationship. Nevertheless, existing counterfactual methods  substitute individual counterfactual observations for individual baseline observations. In contrast, our strategy substitutes counterfactual datasets, with new variable distribution characteristics, for the original dataset.

The first key advantage of this strategy is to preserve as much as possible the distribution of the test set $(X_i^1,\dots ,X_i^p, \hat{Y_i},Y_i ),\, i = 1,\dots, n$ and thus preserving the correlations of the input variables that have then an impact on the indicators computed  by using our procedure. In contrasts with other interpretability paradigms such as the PAC learning framework \citep{valiant2013probably}, we do not create artificial outliers. Its second key advantage is that, for a given perturbation, the weights are obtained by minimizing a convex function, for which the evaluation cost is $\mathcal{O}(n)$. The total cost is then $\mathcal{O}(np)$ for studying the impact of each of the $p$ variables (see  Section~\ref{ssec:ComputaBurden}) and there is no need to generate new data, nor even to compute new predictions from the black box algorithm, which is particularly costly if $n$ or $p$ is large. Our procedure therefore scales particularly well to large datasets as \textit{e.g.} real-world image databases.
Finally, the flexibility of the entropic variable projection procedure enables to study the response to various types of stress on the input variables (not only their mean but also their variability, joint correlations, ...) and thus to interpret the decision rules encountered in a wide range of applications in the field of machine learning. \\
A package in Python with other examples and industrial use cases is available at \url{https://gems-ai.aniti.fr/}.
\vskip .1in

\section*{Funding}
 This work was supported by  the French Investing for the Future   PIA3 program within the Artificial and Natural Intelligence Toulouse Institute (ANITI).

\section*{\textcolor{black}{Acknowledgements}}
\textcolor{black}{The authors are grateful to a Reviewer, whose comments and suggestions enabled to improve the paper.}


%



\appendix

\section{Proofs} \label{s:proofs}

The proofs rely on the following theorem, that is a simplified version  of the Theorems  in \citet{Csis1} and in \citet{Csis2}.

\begin{theorem} \label{th:main}
	Let $(E,\mathcal{B}(E))$ be a measurable space and $Q$ a probability measure on $E$. Consider $t \in \mathbb{R}^k$ and a measurable function $\Phi : E \to \mathbb{R}^k$.
	We assume that, for $v \in \mathbb{R}^k$, $b \in \mathbb{R}$, $Q (\{x \in E ;  \langle v , \Phi(x) \rangle = b \}) = 1$ if and only if $v = 0$ and $b=0$.
	Let $\mathbb{P}_{\Phi,t}$ be the set of all probability measures $P$ on $(E,\mathcal{B}(E))$ such that $
	\int_{E} \Phi(x) \dd P(x)=t$.
	Assume that $\mathbb{P}_{\Phi,t}$ contains a probability measure that is mutually absolutely continuous with respect to $Q$.

	For a  vector $\xi \in \mathbb{R}^k$, let $Z(\xi):=\int_E e^{\langle\xi,\Phi(x) \rangle} \dd Q(x)$. We  assume that the set on which $Z$ is finite is open. Define now $\xi(t)$ as the unique minimizer of the strictly convex function
	$H(\xi):=\log Z(\xi)-\langle\xi,t\rangle$.
	Then,
	\begin{equation}
		Q_t:= \underset{P\in\mathbb{P}_{\Phi,t}}{\mathrm{arginf}} ~ \mathrm{KL}(P,Q) \label{thesol}
	\end{equation}
	exists and is unique. Furthermore it can be computed as
	\begin{align*}
		Q_t
		& =
		\frac{
			\exp \langle \xi(t),  \Phi \rangle
		}
		{
			Z(\xi(t))
		}
		Q.
	\end{align*}
\end{theorem}

{\bf Proof of Theorem \ref{thm:discrete:reweigth:multidim}}
We will apply Theorem \ref{th:main} with $E = \mathbb{R}^{p+2}$ and $Q = Q_n$. Because of the assumption that $t$ can be written as a convex combination of $\Phi(X_1,\hat{Y}_1,Y_1) , \ldots , \Phi(X_n,\hat{Y}_n,Y_n)$, with positive weights, we have that $\mathbb{P}_{\Phi,t}$ in Theorem \ref{th:main} contains a probability measure that is mutually absolutely continuous with respect to $Q$.
Furthermore, we have assumed that the empirical covariance matrix of $ \Phi(X_1,\hat{Y}_1,Y_1) , \ldots , \Phi(X_n,\hat{Y}_n,Y_n) $ is invertible, which means that for any $v \in \mathbb{R}^k$, $b \in \mathbb{R}$, $ \langle v , \Phi(X_1,\hat{Y}_1,Y_1) \rangle ,\ldots , \langle v , \Phi(X_n,\hat{Y}_n,Y_n) \rangle$ are not all equal to $b$. This implies that $Q_n (\{x \in \mathbb{R}^{p+2} ;  \langle v , \Phi(x) \rangle = b \}) = 1$ if and only if $v = 0$ and $b=0$.
Hence, all the assumptions of Theorem \ref{th:main} are satisfied.

We have, starting from the notation of Theorem \ref{th:main},
\[
\int_E e^{\langle\xi,\Phi\rangle}\dd Q(x)
=
\frac{1}{n} \sum_{i=1}^n e^{\langle  \xi , \Phi(X_i,\hat{Y}_i,Y_i)  \rangle}
\]
and thus the definitions of $Z(\xi)$ in Theorems \ref{th:main} and \ref{thm:discrete:reweigth:multidim} indeed coincide. Hence, also the definitions of $\xi(t)$ in Theorems \ref{th:main} and \ref{thm:discrete:reweigth:multidim} coincide.
Hence, we have
\begin{align*}
	Q_t
	& =
	\frac{\exp \langle\xi(t),  \Phi\rangle}{Z(\xi(t))} Q
	\\
	& =
	\frac{1}{ \frac{1}{n} \sum_{i=1}^n e^{\langle \xi(t) , \Phi(X_i,\hat{Y}_i,Y_i)  \rangle} }
	\frac{1}{n}
	\sum_{i=1}^n
	\exp
	\left( \langle\xi(t) ,  \Phi(X_i,\hat{Y_i},Y_i) \rangle
	\right)
	\delta_{X_i,\hat{Y}_i,Y_i}
	\\
	& =
	\frac{1}{n}
	\sum_{i=1}^n
	\lambda_i^{(t)}
	\delta_{X_i , \hat{Y}_i , Y_i}.
\end{align*}
This concludes the proof.
\hfill\ensuremath{\square}

{\bf Proof of Theorem \ref{th:mainenpractice}}
The proof of this theorem comes from Theorem~\ref{thm:discrete:reweigth:multidim}, by considering $\Phi : \mathbb{R}^{p+2} \to \mathbb{R}$ defined by $\Phi(X^1,\ldots,X^p,\hat{Y},Y)=X^{j_0}$ and by considering the same $t \in \mathbb{R}$ in Theorems \ref{thm:discrete:reweigth:multidim} and \ref{th:mainenpractice}.
We have assumed that $\min_{i=1}^n X_i^{j_0} < t < \max_{i=1}^n X_i^{j_0}$. Hence, $t$ can be written as a convex combination of $X_1^{j_0}, \ldots , X_n^{j_0}$ with positive weights.
Furthermore, $X_1^{j_0},\ldots,X_n^{j_0}$ are not all equal and thus their empirical variance is non-zero.
Hence the conditions of Theorem~\ref{thm:discrete:reweigth:multidim} hold and the conclusion of this theorem directly provides Theorem \ref{th:mainenpractice}.
\hfill\ensuremath{\square}

{\bf Proof of Corollary \ref{cor:stress:means}} With the choice of $\Phi$ of the corollary and with the assumptions there, the conditions of Theorem \ref{thm:discrete:reweigth:multidim} hold. Hence the conclusion of this theorem proves the corollary.
\hfill\ensuremath{\square}

{\bf Proof of Corollary \ref{cor:stress:variance}} With the choice of $\Phi$ of the corollary and with the assumptions there, the conditions of Theorem \ref{thm:discrete:reweigth:multidim} hold. Hence the conclusion of this theorem proves the corollary, since $\left( \mathbb{E} (X^{j_0}) = m_{j_0},\mathrm{Var}(X^{j_0}) = v \right)$ is equivalent to $\left( \mathbb{E} (X^{j_0}) = m_{j_0},\mathbb{E}((X^{j_0})^2) = m_{j_0}^2 + v \right)$.
\hfill\ensuremath{\square}

{\bf Proof of Corollary \ref{cor:stress:covariance}} With the choice of $\Phi$ of the corollary and with the assumptions there, the conditions of Theorem \ref{thm:discrete:reweigth:multidim} hold. Hence the conclusion of this theorem proves the corollary, since $\left( \mathbb{E} (X^{j_1}) = m_{j_1},\mathbb{E} (X^{j_2}) = m_{j_2},\mathrm{Cov}(X^{j_1},X^{j_2}) = c \right)$ is equivalent to $ \mathbb{E} (X^{j_1}) = m_{j_1},\mathbb{E} (X^{j_2}) = m_{j_2},\mathbb{E}(X^{j_1}X^{j_2}) = m_{j_1} m_{j_2} + c $.
\hfill\ensuremath{\square}

{\bf Proof of Proposition~\ref{prop:rate:convergence}}
The existence and unicity of $Q_t^\star$ follows from Theorem \ref{th:main}. Also from this theorem,
$Q_t^\star$ is of the form
\[
\dd Q_t^\star (x)
=
e^{ \langle  \xi^{\star}(t) , \Phi(x) \rangle - \log(Z^{\star}(\xi^\star(t))) }
\dd Q^\star(x),
\]
where $\xi^{\star}(t)$ is the minimizer of the strictly convex function
\[
\xi \mapsto
F (\xi) :=
\log(Z^{\star}(\xi)) - \langle \xi , t \rangle
\]
with
\[
Z^\star(\xi) = \int_{\mathbb{R}^{p+2}} e^{\langle \xi , \Phi(x) \rangle} \dd Q^\star(x).
\]
We also recall from Theorem \ref{thm:discrete:reweigth:multidim} that $Q_t$ is of the form
\[
\dd Q_t (x)
=
e^{ \langle  \xi(t) , \Phi(x) \rangle - \log(Z(\xi(t))) }
\dd Q_n(x),
\]
where $\xi(t)$ is the minimizer of the strictly convex function
\[
\xi \mapsto
F_n(\xi) :=
\log(Z(\xi)) - \langle \xi , t \rangle
\]
with
\[
Z(\xi) = \int_{\mathbb{R}^{p+2}} e^{\langle \xi , \Phi(x) \rangle}  \dd Q_n(x).
\]

Let us first prove that $n^{1/2} ( \xi^\star(t) - \xi(t) )  = O_p(1)$. The gradient of $F$ at $\xi$ can be computed as
\[
\nabla_{\xi} F (\xi)
=
\frac{
	\int_{\mathbb{R}^{p+2}} \Phi(x) e^{\langle \xi , \Phi(x) \rangle} \dd Q^\star(x)
}{
	\int_{\mathbb{R}^{p+2}} e^{\langle \xi , \Phi(x) \rangle} \dd Q^\star(x)
}
-t.
\]
The norm of this gradient is bounded by a constant $C_{\sup,1} < \infty$ from the assumption that $\Phi$ is bounded on the support of $Q^{\star}$.

The Hessian matrix of $F$ at $\xi$ is
\begin{align*}
	H_{\xi} F (\xi)
	& =
	\frac{
		\int_{\mathbb{R}^{p+2}} \Phi(x) \Phi(x)^\top e^{\langle \xi , \Phi(x) \rangle} \dd Q^\star(x)
	}{
		\int_{\mathbb{R}^{p+2}} e^{\langle \xi , \Phi(x) \rangle} \dd Q^\star(x)
	}
 -
	\frac{
		\int_{\mathbb{R}^{p+2}} \Phi(x) e^{\langle \xi , \Phi(x) \rangle} \dd Q^\star(x)
	}{
		\int_{\mathbb{R}^{p+2}} e^{\langle \xi , \Phi(x) \rangle} \dd Q^\star(x)
	}
	\frac{
		\int_{\mathbb{R}^{p+2}} \Phi(x)^\top e^{\langle \xi , \Phi(x) \rangle} \dd Q^\star(x)
	}{
		\int_{\mathbb{R}^{p+2}} e^{\langle \xi , \Phi(x) \rangle} \dd Q^\star(x)
	}.
\end{align*}
The two last above fractions are $t$ and $t^\top$ when $\xi = \xi^{\star}(t)$, because $\nabla_{\xi} F (\xi^{\star}(t)) $  is zero because $F$ is minimal at $\xi^{\star}(t)$. Hence, we obtain
\begin{align*}
	H_{\xi} F (\xi^{\star}(t))
	& =
	\frac{
		\int_{\mathbb{R}^{p+2}} \Phi(x) \Phi(x)^\top e^{\langle \xi^{\star}(t) , \Phi(x) \rangle} \dd Q^\star(x)
	}{
		\int_{\mathbb{R}^{p+2}} e^{\langle \xi^{\star}(t) , \Phi(x) \rangle} \dd Q^\star(x)
	}
	-
	t t^\top.
\end{align*}

Hence, $H_{\xi} F (\xi^{\star}(t)) $ is the covariance matrix of $\Phi(\cdot)$, under the probability distribution proportional to $e^{\langle \xi^\star(t) , \Phi( \cdot ) \rangle} \dd Q^\star( \cdot )$ on $\mathbb{R}^{p+2}$.
Assume that $H_{\xi} F (\xi) $ is not invertible. Then there exists an affine subspace $G$ of $\mathbb{R}^k$ with dimension strictly less than $k$ that contains $\Phi(\cdot)$ almost surely under the probability distribution proportional to $e^{\langle \xi^\star(t) , \Phi( \cdot ) \rangle} \dd Q^\star( \cdot )$. Since $\Phi$ is bounded on the support of $Q^{\star}$, this distribution is equivalent to $\dd Q^\star( \cdot )$ and thus $\phi(\cdot) \in G$ almost surely under the probability distribution $\dd Q^\star( \cdot )$. This is in contradiction with an assumption of Proposition~\ref{prop:rate:convergence}. Hence, $H_{\xi} F (\xi^{\star}(t)) $ is invertible.

Let $B(\xi, r)$ be the Euclidean ball with center $\xi$ and radius $r$.
One can see that the partial derivatives of $H_{\xi}F(\xi)$ are bounded on $B(\xi^\star(t) , \delta)$ using that $\Phi$ is bounded on the support of $Q^{\star}$. Hence, there are constants $C_{\inf,2} >0$ and $\delta >0$ such that
\begin{equation} \label{eq:lambda:inf:large}
	\inf_{\xi \in B(\xi^\star(t) , \delta)}
	\lambda_{\inf} (  H_{\xi} F (\xi^{\star}(t))   )
	\geq
	C_{\inf,2},
\end{equation}
with $\lambda_{\inf} ( \cdot )$ the smallest eigenvalue.

Furthermore, we can compute similarly the Hessian matrix of $F_n$ at $\xi$,
\begin{align*}
	H_{\xi} F_n (\xi)
	& =
	\frac{
		\int_{\mathbb{R}^{p+2}} \Phi(x) \Phi(x)^\top e^{\langle \xi , \Phi(x) \rangle} \dd Q_n(x)
	}{
		\int_{\mathbb{R}^{p+2}} e^{\langle \xi , \Phi(x) \rangle} \dd Q_n(x)
	}
 -
	\frac{
		\int_{\mathbb{R}^{p+2}} \Phi(x) e^{\langle \xi , \Phi(x) \rangle} \dd Q_n(x)
	}{
		\int_{\mathbb{R}^{p+2}} e^{\langle \xi , \Phi(x) \rangle} \dd Q_n(x)
	}
	\frac{
		\int_{\mathbb{R}^{p+2}} \Phi(x)^\top e^{\langle \xi , \Phi(x) \rangle} \dd Q_n(x)
	}{
		\int_{\mathbb{R}^{p+2}} e^{\langle \xi , \Phi(x) \rangle} \dd Q_n(x)
	}.
\end{align*}
Hence, one can see that the partial derivatives of $H_{\xi} F_n (\xi) $ are bounded uniformly in $\xi \in B(\xi^\star(t) , \delta)$, with a fixed deterministic bound, using again that $\Phi$ is bounded on the support of $Q^{\star}$. Furthermore, for any fixed $\xi$, from the law of large number $H_{\xi} F_n (\xi) \to H_{\xi} F (\xi) $ almost surely. Hence, we can show
\begin{equation} \label{eq:converge:hessian}
	\sup_{\xi \in B(\xi^\star(t) , \delta ) }
	\left|
	\lambda_{\inf}
	(
	H_{\xi} F_n (\xi)
	)
	-
	\lambda_{\inf}
	(
	H_{\xi} F (\xi)
	)
	\right|
	=
	o_p(1).
\end{equation}

Let us consider $M >0$ to be fixed later. By convexity, we have
\begin{align} \label{eq:proba:to:bound}
	&
	\mathbb{P}
	\left(
	|| \xi(t) - \xi^\star(t) ||
	\geq
	\frac{M}{\sqrt{n}}
	\right)
	\leq
	\notag
	\\
	&
	\mathbb{P}
	\left(
	\inf_{ || \xi - \xi^{\star}(t) || = M / n^{1/2} }
	\left(
	F_n(\xi) - F_n(\xi^\star(t))
	\right)
	\leq 0
	\right).
\end{align}
Let us bound the probability of this last event. We have, for some random $\hat{\xi} \in B( \xi^{\star}(t)  , M/n^{1/2} )$ and $\hat{\hat{\xi}}$ with $||\hat{\hat{\xi}} - \xi^{\star}(t) || = M/n^{1/2}$,
\begin{align*}
	& \inf_{ || \xi - \xi^{\star}(t) || = M / n^{1/2} }
	\left(
	F_n(\xi) - F_n(\xi^\star(t))
	\right)
 =
	(\nabla_{\xi} F_n( \xi^\star(t) ))^\top
	( \hat{\hat{\xi}}  -  \xi^{\star}(t))
	+
	\frac{1}{2}
	( \hat{\hat{\xi}}  -  \xi^{\star}(t))^\top
	H_\xi F_n( \hat{\xi} )
	( \hat{\hat{\xi}}  -  \xi^{\star}(t)).
\end{align*}
Since $\nabla_{\xi} F( \xi^\star(t) ) = 0$ we can show simply $|| \nabla_{\xi} F_n( \xi^\star(t) ) || = O_p(n^{-1/2})$. Furthermore, from \eqref{eq:converge:hessian} and \eqref{eq:lambda:inf:large}, we obtain
\begin{align*}
	& \inf_{ || \xi - \xi^{\star}(t) || = M / n^{1/2} }
	\left(
	F_n(\xi) - F_n(\xi^\star(t))
	\right)
	 \geq
	- O_p \left( \frac{1}{\sqrt{n}}  \right)  \frac{M}{\sqrt{n}}
	+
	\frac{1}{2}
	\left(
	\frac{M}{\sqrt{n}}
	\right)^2
	\left(
	C_{\inf,2} + o_p(1)
	\right).
\end{align*}
The above $O_p(1)$ and $o_p (1)$ can be taken to be independent on $M$. Hence, for any $\eta >0$ we can take $M$ large enough such that \eqref{eq:proba:to:bound} is smaller than $\eta$. Hence we have shown
\begin{equation} \label{eq:xi:star:minus:xi:rate}
	|| \xi^\star(t) - \xi(t) ||
	=
	O_p
	\left(
	\frac{1}{\sqrt{n}}
	\right).
\end{equation}

We have
\begin{align}
	Z^\star( \xi^\star(t) )
	-
	Z(\xi(t))
	\notag
 & =
	\int_{\mathbb{R}^{p+2}}
	e^{\langle \Phi(x) , \xi^\star(t) \rangle}
	\dd Q^\star
	-
	\int_{\mathbb{R}^{p+2}}
	e^{\langle \Phi(x) , \xi(t) \rangle}
	\dd Q_n
	\notag
	\\
	& =
	\int_{\mathbb{R}^{p+2}}
	e^{\langle \Phi(x) , \xi^\star(t) \rangle}
	\dd Q^\star
	-
	\int_{\mathbb{R}^{p+2}}
	e^{\langle \Phi(x) , \xi^\star(t) \rangle}
	\dd Q_n
	\label{eq:Z:star:one}
	\\
	& +
	\int_{\mathbb{R}^{p+2}}
	e^{\langle \Phi(x) , \xi^\star(t) \rangle}
	\dd Q_n
	-
	\int_{\mathbb{R}^{p+2}}
	e^{\langle \Phi(x) , \xi(t) \rangle}
	\dd Q_n.
	\label{eq:Z:star:two}
\end{align}
The quantity in \eqref{eq:Z:star:one} is a $O_p(n^{-1/2})$ by the central limit theorem since $\Phi$ is bounded on the support of $Q^{\star}$. The quantity in \eqref{eq:Z:star:two} is a $O_p(n^{-1/2})$ from \eqref{eq:xi:star:minus:xi:rate} and because $\Phi$ is bounded on the support of $Q^{\star}$. Hence, we have shown
\begin{equation} \label{eq:Z:star:minus:Z:rate}
	Z^\star( \xi^\star(t) )
	-
	Z(\xi(t))
	=
	O_p \left(
	\frac{1}{\sqrt{n}}
	\right).
\end{equation}

Let now $\mathcal{F} = \{ f: \mathbb{R}^{p+2} \to \mathbb{R} ; \text{f is $1$-Lipschitz} , f(0) = 0 \}$. We have
\begin{align}
	&
	\sup_{f \in \mathcal{F}}
	\left|
	\int_{\mathbb{R}^{p+2}}
	f(x) \dd Q^\star_t (x)
	-
	\int_{\mathbb{R}^{p+2}}
	f(x) \dd Q_t (x)
	\right|
\notag
	\\
	& 
 \leq
	\frac{1}{Z^{\star}( \xi^\star(t) )}
	\sup_{f \in \mathcal{F}}
	\Big|
	\int_{\mathbb{R}^{p+2}}
	f(x)
	e^{ \langle  \xi^\star(t) , \Phi(x) \rangle  }
	\left(
	\dd Q^\star (x)
	-
	\dd Q_n(x)
	\right)
	\Big| \label{eq:proof:rate:one}
	\\
	& +
	\frac{1}{Z^{\star}( \xi^\star(t) )}
	\sup_{f \in \mathcal{F}}
	\Big|
	\int_{\mathbb{R}^{p+2}}
	f(x)
	e^{ \langle  \xi^{\star}(t) , \Phi(x) \rangle }
	\dd Q_n(x)
	-
	\int_{\mathbb{R}^{p+2}}
	f(x)
	e^{ \langle  \xi(t) , \Phi(x) \rangle }
	\dd Q_n(x)
	\Big|
	\label{eq:proof:rate:two}
	\\
	&  +
	\left|
	\frac{1}{Z^{\star}( \xi^\star(t) )}
	-
	\frac{1}{Z( \xi(t) )}
	\right|
	\sup_{f \in \mathcal{F}}
	\left|
	\int_{\mathbb{R}^{p+2}}
	f(x)
	e^{ \langle  \xi(t) , \Phi(x) \rangle }
	\dd Q_n(x)
	\right|.
	\label{eq:proof:rate:three}
\end{align}

The term \eqref{eq:proof:rate:two} is smaller than a constant times $||\xi^\star(t) - \xi(t)||$ because $\Phi$ is bounded on the support of $Q^\star$ and $f$ is bounded. Hence this term is a $O_p(n^{-1/2})$ from \eqref{eq:xi:star:minus:xi:rate}.
The term in  \eqref{eq:proof:rate:three} is also a $O_p(n^{-1/2})$ from \eqref{eq:Z:star:minus:Z:rate}. Let us finally address the term in \eqref{eq:proof:rate:one}. In this term, the function that is integrated is uniformly bounded, with uniformly bounded Lipschitz norm, because $\Phi$ is bounded and Lipschitz continuous on the bounded support of $Q^\star$ and since $f \in \mathcal{F}$. Also, from for instance Theorem 1 in \citet{fournier2015rate}, the $L^1$ Wasserstein distance between $Q_n$ and $Q^\star$
satisfies
\[
\mathcal{W}_1
\left(
Q_n,Q^{\star}
\right)
= O_p
\left(
n^{-1/(p+2)}
\right).
\]

This implies that the supremum over $f \in \mathcal{F}$ in \eqref{eq:proof:rate:one} is bounded by a constant times $O_p
\left(
n^{-1/(p+2)}
\right)$,
see for instance \citet{villani2008optimal} for the link between the $L^1$ Wasserstein distance and differences of expectations of Lipschitz functions.  Hence we have proved that as $n \to \infty$,
\begin{flalign*}
	&
	\sup_{f \in \mathcal{F}}
	\left|
	\int_{\mathbb{R}^{p+2}}
	f(x) \dd Q^\star_t (x)
	-
	\int_{\mathbb{R}^{p+2}}
	f(x) \dd Q_t (x)
	\right|
	= O_p \left(
	n^{-1/(p+2)}
	\right).
\end{flalign*}
From for instance \citet{villani2008optimal}, this implies the conclusion of the proposition.
\hfill\ensuremath{\square}



\begin{thebibliography}{}
	
	\bibitem[\protect\citename{Bachoc {\em et~al.}, }2017]{bachoc2017gaussian}
	Bachoc, Fran{\c{c}}ois, Gamboa, Fabrice, Loubes, Jean-Michel, \& Venet, Nil.
	2017.
	\newblock A Gaussian process regression model for distribution inputs.
	\newblock {\em IEEE Transactions on Information Theory}, {\bf 64}(10),
	6620--6637.
	
	\bibitem[\protect\citename{Baehrens {\em et~al.}, }2010]{baehrens2010explain}
	Baehrens, David, Schroeter, Timon, Harmeling, Stefan, Kawanabe, Motoaki,
	Hansen, Katja, \& M\"uller, Klaus-Robert. 2010.
	\newblock How to explain individual classification decisions.
	\newblock {\em Journal of Machine Learning Research}, {\bf 11}(Jun),
	1803--1831.
	
	\bibitem[\protect\citename{Black {\em et~al.}, }2020]{FlipTest2020}
	Black, Emily, Yeom, Samuel, \& Fredrikson, Matt. 2020.
	\newblock FlipTest: fairness testing via optimal transport.
	\newblock {\em Pages  111--121 of:} {\em FAT* '20: Proceedings of the 2020
		Conference on Fairness, Accountability, and Transparency}.
	
	\bibitem[\protect\citename{Breiman, }2001]{breiman2001random}
	Breiman, Leo. 2001.
	\newblock Random forests.
	\newblock {\em Machine Learning}, {\bf 45}(1), 5--32.
	
	
		\bibitem[\protect\citename{Broto {\em et~al.}, }2020]{broto2020variance}
	Broto, Baptiste, Bachoc, Fran{\c{c}}ois \& Depecker, Marine. 2020.
	\newblock Variance reduction for estimation of Shapley effects and adaptation to unknown input distribution.
	\newblock {\em SIAM/ASA Journal on Uncertainty Quantification}, {\bf 8}(2), 693--716.

	
	\bibitem[\protect\citename{B{\"u}hlmann \& Van De~Geer,
	}2011]{buhlmann2011introduction}
	B{\"u}hlmann, Peter, \& Van De~Geer, Sara. 2011.
	\newblock {\em Statistics for High-Dimensional Data}.
	\newblock Springer.
	
	\bibitem[\protect\citename{Caruana {\em et~al.},
	}2015]{Caruana:2015:IMH:2783258.2788613}
	Caruana, Rich, Lou, Yin, Gehrke, Johannes, Koch, Paul, Sturm, Marc, \& Elhadad,
	Noemie. 2015.
	\newblock Intelligible models for healthcare: predicting pneumonia risk and
	hospital 30-day readmission.
	\newblock {\em Pages  1721--1730 of:} {\em Proceedings of the 21th ACM SIGKDD
		International Conference on Knowledge Discovery and Data Mining}.
	\newblock KDD '15.
	
	\bibitem[\protect\citename{Cazelles {\em et~al.}, }2018]{cazelles18geodesic}
	Cazelles, Elsa, Seguy, Vivien, Bigot, Jérémie, Cuturi, Marco, \& Papadakis,
	Nicolas. 2018.
	\newblock Geodesic PCA versus Log-PCA of histograms in the {Wasserstein} space.
	\newblock {\em SIAM Journal on Scientific Computing}, {\bf 40}(2), B429--B456.
	
	\bibitem[\protect\citename{Craven \& Shavlik, }1995]{DBLP:conf/nips/CravenS95}
	Craven, Mark, \& Shavlik, Jude~W. 1995.
	\newblock Extracting tree-structured representations of trained networks.
	\newblock {\em Pages  24--30 of:} {\em Advances in Neural Information
		Processing Systems 8, NIPS, Denver, CO, November 27-30, 1995}.
	
	\bibitem[\protect\citename{Csisz{\'a}r, }1975]{Csis1}
	Csisz{\'a}r, Imre. 1975.
	\newblock \mbox{$I$}-divergence geometry of probability distributions and
	minimization problems.
	\newblock {\em The Annals of Probability},  146--158.
	
	\bibitem[\protect\citename{Csisz{\'a}r, }1984]{Csis2}
	Csisz{\'a}r, Imre. 1984.
	\newblock Sanov property, generalized \mbox{$I$}-projection and a conditional
	limit theorem.
	\newblock {\em The Annals of Probability},  768--793.
	
\bibitem[\protect\citename{De Lara {\em et~al.}, }2021]{de2021transport}
De Lara, Lucas and Gonz{a}lez-Sanz, Alberto and Asher, Nicholas and Loubes, Jean-Michel. 2021.
\newblock{Transport-based counterfactual models},
\newblock{\em  arXiv preprint arXiv:2108.13025}.
	
	\bibitem[\protect\citename{Dzindolet {\em et~al.}, }2003]{dzindolet2003role}
	Dzindolet, Mary~T, Peterson, Scott~A, Pomranky, Regina~A, Pierce, Linda~G, \&
	Beck, Hall~P. 2003.
	\newblock The role of trust in automation reliance.
	\newblock {\em International Journal of Human-Computer Studies}, {\bf 58}(6),
	697--718.
	
	\bibitem[\protect\citename{Fournier \& Guillin, }2015]{fournier2015rate}
	Fournier, Nicolas, \& Guillin, Arnaud. 2015.
	\newblock On the rate of convergence in {Wasserstein} distance of the empirical
	measure.
	\newblock {\em Probability Theory and Related Fields}, {\bf 162}(3-4),
	707--738.
	
	\bibitem[\protect\citename{Goyal {\em et~al.}, }2019]{Goyal_ICML2019}
	Goyal, Yash, Wu, Ziyan, Ernst, Jan, Batra, Dhruv, Parikh, Devi, \& Lee, Stefan.
	2019.
	\newblock Counterfactual visual explanations.
	\newblock {\em Pages  2376--2384 of:} {\em Proceedings of the 36th
		International Conference on Machine Learning}.
	
	\bibitem[\protect\citename{Hall {\em et~al.}, }2018]{hall2018practical}
	Hall, Patrick, Gill, Navdeep, \& Chan, Mark. 2018.
	\newblock {\em Practical Techniques for Interpreting Machine Learning Models:
		Introductory Open Source Examples Using Python, H2O, and XGBoost}.
	
	\bibitem[\protect\citename{Hastie {\em et~al.}, }2009]{trevor2009elements}
	Hastie, Trevor, Tibshirani, Robert, \& Friedman, Jerome~H. 2009.
	\newblock {\em The Elements of Statistical Learning: Data Mining, Inference,
		and Prediction}.
	\newblock New York, NY: Springer.
	
	\bibitem[\protect\citename{He {\em et~al.}, }2016]{He2016DeepRL}
	He, Kaiming, Zhang, Xiangyu, Ren, Shaoqing, \& Sun, Jian. 2016.
	\newblock Deep residual learning for image recognition.
	\newblock {\em 2016 IEEE Conference on Computer Vision and Pattern Recognition
		(CVPR)},  770--778.
	
	\bibitem[\protect\citename{Herlocker {\em et~al.},
	}2000]{herlocker2000explaining}
	Herlocker, Jonathan, Konstan, Joseph, \& Riedl, John. 2000.
	\newblock Explaining collaborative filtering recommendations.
	\newblock {\em Pages  241--250 of:} {\em Proceedings of the 2000 ACM conference
		on Computer supported cooperative work}.
	\newblock ACM.
	
	\bibitem[\protect\citename{Hooker {\em et~al.}, }2019]{Hooker2019ABF}
	Hooker, Sara, Erhan, Dumitru, Kindermans, Pieter-Jan, \& Kim, Been. 2019.
	\newblock A benchmark for interpretability methods in deep neural networks.
	\newblock {\em Pages  9737--9748 of:} {\em Advances in Neural Information
		Processing Systems}.
	
	\bibitem[\protect\citename{Ignatiev {\em et~al.}, }2019]{NIPS2019_9717}
	Ignatiev, Alexey, Narodytska, Nina, \& Marques-Silva, Joao. 2019.
	\newblock On relating explanations and adversarial examples.
	\newblock {\em Pages  15857--15867 of:} {\em Advances in Neural Information
		Processing Systems 32}.
	
	\bibitem[\protect\citename{Koh \& Liang, }2017]{KohMLR2017}
	Koh, Pang~Wei, \& Liang, Percy. 2017.
	\newblock Understanding black-box predictions via influence functions.
	\newblock {\em In:} {\em Proceedings of the 34th International Conference on
		Machine Learning (ICML 2017)}.
	
	\bibitem[\protect\citename{Kramer \& Choudhary, }2018]{skater}
	Kramer, Aaron, \& Choudhary, Pramit. 2018.
	\newblock {\em {Model Interpretation with Skater}}.
	\newblock \url{https://oracle.github.io/Skater/}.
	\newblock [Online; accessed 28-Jan-2020].
	
	\bibitem[\protect\citename{Kusner {\em et~al.}, }2017]{NIPS2017_6995}
	Kusner, Matt~J, Loftus, Joshua, Russell, Chris, \& Silva, Ricardo. 2017.
	\newblock Counterfactual fairness.
	\newblock {\em Pages  4066--4076 of:} {\em Advances in Neural Information
		Processing Systems 30}.
	
	\bibitem[\protect\citename{Lema{\^\i}tre {\em et~al.},
	}2015]{lemaitre2015density}
	Lema{\^\i}tre, Paul, Sergienko, Ekatarina, Arnaud, Aur{\'e}lie, Bousquet,
	Nicolas, Gamboa, Fabrice, \& Iooss, Bertrand. 2015.
	\newblock Density modification-based reliability sensitivity analysis.
	\newblock {\em Journal of Statistical Computation and Simulation}, {\bf 85}(6),
	1200--1223.
	
	\bibitem[\protect\citename{Lipton, }2016]{LiptonICML2016}
	Lipton, Zachary~C. 2016.
	\newblock The mythos of model interpretability.
	\newblock {\em Pages  96--100 of:} {\em Proceedings of the ICML Workshop on
		Human Interpretability in Machine Learning (WHI 2016)}.
	
	\bibitem[\protect\citename{Liu {\em et~al.}, }2015]{liu2015faceattributes}
	Liu, Ziwei, Luo, Ping, Wang, Xiaogang, \& Tang, Xiaoou. 2015 (December).
	\newblock Deep learning face attributes in the wild.
	\newblock {\em In:} {\em Proceedings of International Conference on Computer
		Vision (ICCV)}.
	
	\bibitem[\protect\citename{Lou {\em et~al.},
	}2012]{Lou:2012:IMC:2339530.2339556}
	Lou, Yin, Caruana, Rich, \& Gehrke, Johannes. 2012.
	\newblock Intelligible models for classification and regression.
	\newblock {\em Pages  150--158 of:} {\em Proceedings of the 18th ACM SIGKDD
		International Conference on Knowledge Discovery and Data Mining}.
	\newblock KDD '12.
	
	\bibitem[\protect\citename{Lundberg \& Lee, }2017]{shap}
	Lundberg, Scott~M, \& Lee, Su-In. 2017.
	\newblock A unified approach to interpreting model predictions.
	\newblock {\em Pages  4765--4774 of:} {\em Advances in Neural Information
		Processing Systems (NIPS)}.
	
	\bibitem[\protect\citename{Montavon {\em et~al.}, }2017]{montavon2017methods}
	Montavon, Gr{\'e}goire, Samek, Wojciech, \& M{\"u}ller, Klaus-Robert. 2017.
	\newblock Methods for interpreting and understanding deep neural networks.
	\newblock {\em Digital Signal Processing}.


	\bibitem[\protect\citename{Owen \& Prieur, }2017]{owen_shapley_2017}
	Owen, Art B. \& Prieur, Clémentine 2017.
	\newblock On {Shapley} value for measuring importance of dependent inputs.
	\newblock {\em SIAM/ASA Journal on Uncertainty Quantification},
	{\bf 5}(1), 986--1002.
	
	\bibitem[\protect\citename{Pan \& Yang, }2009]{pan2009survey}
	Pan, Sinno, Jialin \& Yang, Qiang 2009.
	\newblock A survey on transfer learning.
	\newblock {\em IEEE Transactions on Knowledge and Data Engineering},
	{\bf 22}(10), 1345--1359.
	
	\bibitem[\protect\citename{Peyr{\'e} \& Cuturi, }2019]{peyre2019computational}
	Peyr{\'e}, Gabriel, \& Cuturi, Marco. 2019.
	\newblock Computational Optimal Transport: With Applications to Data Science.
	\newblock {\em Foundations and Trends{\textregistered} in Machine Learning},
	{\bf 11}(5-6), 355--607.
	
	\bibitem[\protect\citename{Raileanu \& Stoffel, }2004]{Raileanu2004}
	Raileanu, Laura~Elena, \& Stoffel, Kilian. 2004.
	\newblock Theoretical comparison between the Gini index and information gain
	criteria.
	\newblock {\em Annals of Mathematics and Artificial Intelligence}, {\bf 41}(1),
	77--93.
	
	\bibitem[\protect\citename{Ribeiro {\em et~al.}, }2016]{ribeiro2016should}
	Ribeiro, Marco~Tulio, Singh, Sameer, \& Guestrin, Carlos. 2016.
	\newblock {Why should I trust you?: explaining the predictions of any
		classifier}.
	\newblock {\em Pages  1135--1144 of:} {\em Proceedings of the 22nd ACM SIGKDD
		International Conference on Knowledge Discovery and Data Mining}.
	\newblock ACM.
	
	\bibitem[\protect\citename{Saltelli {\em et~al.}, }2008]{saltelli2008global}
	Saltelli, Andrea, Ratto, Marco, Andres, Terry, Campolongo, Francesca, Cariboni,
	Jessica, Gatelli, Debora, Saisana, Michaela, \& Tarantola, Stefano. 2008.
	\newblock {\em Global Sensitivity Analysis: the Primer}.
	\newblock John Wiley \& Sons.
	
	\bibitem[\protect\citename{Selvaraju {\em et~al.}, }2016]{selvaraju2016grad}
	Selvaraju, Ramprasaath, Das, Abhishek, Vedantam, Ramakrishna, Cogswell,
	Michael, Parikh, Devi, \& Batra, Dhruv. 2016.
	\newblock Grad-CAM: why did you say that?
	\newblock {\em arXiv preprint arXiv:1611.07450}.
	

	\bibitem[\protect\citename{Shapley, }1953]{shapley_value_1953}
	Shapley, Lloyd.
	\newblock A value for n-person games. {Contribution} to the {theory} of {games}.
	\newblock {\em Annals of Mathematics Studies}, {\bf 2}, 28.
	
	\bibitem[\protect\citename{Sundararajan {\em et~al.}, }2017]{MukundICML17}
	Sundararajan, Mukund, Taly, Ankur, \& Yan, Qiqi. 2017.
	\newblock Axiomatic attribution for deep networks.
	\newblock {\em Page  3319–3328 of:} {\em Proceedings of the 34th
		International Conference on Machine Learning - Volume 70}.
	\newblock ICML’17.
	
	\bibitem[\protect\citename{Valiant, }2013]{valiant2013probably}
	Valiant, Leslie. 2013.
	\newblock {\em Probably Approximately Correct: Nature's Algorithms for Learning
		and Prospering in a Complex World}.
	\newblock Basic Books (AZ).
	
	\bibitem[\protect\citename{Villani, }2008]{villani2008optimal}
	Villani, C{\'e}dric. 2008.
	\newblock {\em Optimal Transport: Old and New}.
	\newblock  Vol. 338.
	\newblock Springer Science \& Business Media.
	
	\bibitem[\protect\citename{Wachter {\em et~al.}, }2018]{WachterHJLT18}
	Wachter, Sandra, Mittelstadt, Brent, \& Russell, Chris. 2018.
	\newblock Counterfactual explanations without opening the black box: automated
	decisions and the GDPR.
	\newblock {\em Harvard Journal of Law \& Technology}, {\bf 31}(04), 841--887.
	
		\bibitem[\protect\citename{Wäldchen {\em et~al.}, }2021]{waldchen2021}
	Wäldchen, Stephan, MacDonald, Jan, Hauch, Sascha and Kutyniok, Gitta. 2021.
	\newblock The computational complexity of understanding binary classifier decisions.
	\newblock {\em Journal of Artificial Intelligence Research}, {\bf 70}, 351--387.
	

	
\end{thebibliography}
\end{document}